%% file: final.tex
\documentclass[journal,twoside,web]{ieeecolor}
\usepackage{tmi}
\usepackage{cite}
\usepackage{amsmath,amssymb,amsfonts}
\usepackage{algorithmic}
\usepackage{graphicx}
\usepackage{multirow}
\usepackage{textcomp}
\usepackage{color}
\usepackage{subcaption}
\usepackage[nice]{nicefrac}
\usepackage[normalem]{ulem}

\usepackage{amsmath}
\DeclareMathOperator*{\argmax}{argmax}

\makeatletter
\let\NAT@parse\undefined
\makeatother
\usepackage{hyperref}

\def\BibTeX{{\rm B\kern-.05em{\sc i\kern-.025em b}\kern-.08em
    T\kern-.1667em\lower.7ex\hbox{E}\kern-.125emX}}
\begin{document}
\title{Dual Cross-image Semantic Consistency \\with Self-aware Pseudo Labeling for Semi-\\supervised Medical Image Segmentation}

\author{
{
Han Wu, 
Chong Wang, \textit{Member, IEEE},
Zhiming Cui
}
\thanks{
This work was supported by the National Natural Science Foundation of China under Grant 6230012077 and the Shanghai Municipal Central Guided Local Science and Technology Development Fund Project (No. YDZX20233100001001). 
\textit{(Han Wu and Chong Wang contributed equally to this work.)}
\textit{(Corresponding author: Zhiming Cui.)}
}
\thanks{
Han Wu is with the School of Biomedical Engineering \& State Key Laboratory of Advanced Medical Materials and Devices, ShanghaiTech University, Shanghai, 201210, China. He is also with Lingang Laboratory, Shanghai, 200031, China. (e-mail: wuhan2022@shanghaitech.edu.cn).
}
\thanks{
Chong Wang is with the Department of Radiology, Stanford University, Stanford, CA 94305-5105, USA (e-mail: chongwa@stanford.edu)
}
\thanks{
Zhiming Cui is with the School of Biomedical Engineering \& State Key Laboratory of Advanced Medical Materials and Devices, ShanghaiTech University, Shanghai, 201210, China.
(e-mail: cuizhm@shanghaitech.edu.cn).
}
}

\maketitle

\begin{abstract}
Semi-supervised learning has proven highly effective in tackling the challenge of limited labeled training data in medical image segmentation. 
In general, current approaches, which rely on intra-image pixel-wise consistency training via pseudo-labeling, overlook the consistency at more comprehensive semantic levels (e.g., object region) and suffer from severe discrepancy of extracted features resulting from an imbalanced number of labeled and unlabeled data. 
To overcome these limitations, we present a new \underline{Du}al \underline{C}ross-\underline{i}mage \underline{S}emantic \underline{C}onsistency (DuCiSC) learning framework, for semi-supervised medical image segmentation. 
Concretely, beyond enforcing pixel-wise semantic consistency, 
DuCiSC proposes dual paradigms to encourage region-level semantic consistency across: 1) labeled and unlabeled images; and 2) labeled and fused images, 
by explicitly aligning their prototypes. 
Relying on the dual paradigms, 
DuCiSC can effectively establish consistent cross-image semantics via prototype representations, 
thereby addressing the feature discrepancy issue. 
Moreover, we devise a novel self-aware confidence estimation strategy to accurately select reliable pseudo labels, allowing for exploiting the training dynamics of unlabeled data. 
Our DuCiSC method is extensively validated on four datasets, including two popular binary benchmarks in segmenting the left atrium and pancreas, a multi-class Automatic Cardiac Diagnosis Challenge dataset, and
a challenging scenario of segmenting the inferior alveolar nerve that features complicated anatomical structures,
showing superior segmentation results over previous state-of-the-art approaches. 
Our code is publicly available at \href{https://github.com/ShanghaiTech-IMPACT/DuCiSC}{https://github.com/ShanghaiTech-IMPACT/DuCiSC}.

\end{abstract}


\section{Introduction}
\label{sec:introduction}
\IEEEPARstart{A}{ccurate} segmentation of medical images serves as a preliminary and crucial step for computer-assisted diagnosis applications~\cite{fang2019attention,wang2022bowelnet}. 
Recent advancements in deep learning networks have greatly enhanced the field~\cite{zhao2025clip,wang2025mixture},
yet these techniques typically require a substantial amount of pixel-wise or voxel-wise annotated training samples, which are costly and time-consuming to acquire. 
Therefore, alternative semi-supervised approaches that employ a small labeled set alongside a large unlabeled set have been explored, offering a promising solution to reduce the annotation burden and improve the segmentation performance~\cite{wang2023interpretable}.

State-of-the-art (SOTA) approaches in semi-supervised segmentation commonly rely on two pivotal techniques: 
1) pseudo labeling, which creates confident labels for unlabeled samples that are employed to re-train the segmentation model~\cite{zhang2022boostmis}; and 
2) consistency regularization, which enforces consistent model outputs under different forms of perturbations (e.g., input or feature)~\cite{laine2016temporal}. 
One of the most successful approaches is the Mean-teacher (MT) framework~\cite{tarvainen2017mean,cui2019semi}, which 
combines these two techniques by averaging the network parameters during training, yielding high-quality pseudo labels for the unlabeled data to regularize the model's prediction consistency. 
The effectiveness and simplicity of the MT strategy have motivated the development of its many advanced variants for the semi-supervised medical image segmentation~\cite{yu2019uncertainty,you2022simcvd,xu2023ambiguity,bai2023bidirectional}. 

% its extensions~\cite{yu2019uncertainty,you2022simcvd,xu2023ambiguity,bai2023bidirectional}  

% Recently, several studies~\cite{wu2021semi,wu2022mutual,bai2023bidirectional} have endeavored to integrate both approaches to achieve a better segmentation performance. 
% For example, \cite{yu2019uncertainty,bai2023bidirectional} generates the pseudo label for the unlabeled data by the teacher network based on Meat-Teacher~\cite{tarvainen2017mean,cui2019semi}. 
% \cite{wu2021semi,wu2022mutual,su2024mutual} employ dual decoders, where the predictions from each decoder mutually serve as pseudo labels for the other.
% The quality of such pseudo labels significantly impacts segmentation results; 
% however, they can be unstable in the initial stages of training due to the lack of sufficient annotated data, leading to suboptimal performance. 
% Therefore, it is crucial to identify the most confident areas from the pseudo labels for effective training~\cite{yu2019uncertainty,xu2023ambiguity}.

Despite their achievements, these existing approaches still face imperfections. 
Firstly, they mostly focus on the consistency regularization of model outputs only at the voxel level~\cite{yu2019uncertainty,xu2022all,wu2022mutual,xu2023ambiguity,bai2023bidirectional,su2024mutual}, 
while neglecting the consistency at more comprehensive semantic levels, e.g., object region.
Secondly, 
% since the labeled data constitute only a small fraction of the entire dataset, 
they harness the labeled and unlabeled data under a separate learning scheme (e.g., ground-truth labels for labeled samples and pseudo labels for unlabeled samples), 
often leading to a significant discrepancy between the features extracted from labeled and unlabeled training data. 
This phenomenon, known as the empirical distribution mismatch~\cite{bai2023bidirectional}, can severely hinder the model's generalization capacity. 
% It’s hard to use few labeled data to construct the precise distribution of the whole dataset. (d) By using our BCP, the empirical distributions of labeled and unlabeled features are aligned.
% BCP~\cite{bai2023bidirectional} introduced a bi-directional copy-paste strategy based on mixups to tackle this problem. However, the structural information, which is significant for challenging segmentation tasks, i.e. the inferior alveolar nerve, can be compromised during the copy-paste process. 
Lastly, to select reliable pseudo-labeled samples (i.e., pixels or voxels in segmentation tasks) for the model's learning, 
some approaches use predefined fixed confidence thresholds~\cite{mahmood2024splal}, which fail to dynamically reflect the model's learning status. 
Other approaches rely on the model's output, such as calculating entropy~\cite{yu2019uncertainty,xu2023ambiguity}, 
but could still incur confirmation bias~\cite{arazo2020pseudo}, where the supervision of incorrect pseudo labels will increase the model confidence in inaccurate predictions and consequently decrease the accuracy.

In this paper, we present an effective \textbf{Du}al \textbf{C}ross-\textbf{i}mage \textbf{S}emantic \textbf{C}onsistency (DuCiSC) learning framework, based on the MT strategy, for the task of semi-supervised segmentation in medical images. 
Apart from ensuring the pixel-level consistency, 
DuCiSC leverages dual paradigms to complementarily enforce the consistency of region-level semantics characterized by the representation of object (e.g., organ) prototypes. 
To be specific, DuCiSC explicitly aligns prototypes extracted from two pairs of training images: 
1) labeled and unlabeled images; and 
2) labeled and fused images. 
Relying on these dual paradigms, 
DuCiSC can effectively establish consistent prototype representations of cross-image semantics, 
thereby addressing the distribution mismatch issue mentioned earlier. 
In addition, we propose a generalized and self-aware confidence estimation strategy to accurately select reliable pseudo labels, enabling DuCiSC to take advantage of the training dynamics of unlabeled data. 
We extensively validate our DuCiSC method on popular datasets, including three popular benchmarks for medical image segmentation: left atrium (LA), pancreas (NIH-Pancreas), and ACDC.
Additionally, we further included a new challenging scenario of segmenting inferior alveolar nerves that have complicated anatomical structures. 
In summary, the major contributions of this paper are listed as follows:
\begin{enumerate}
    \item We present the DuCiSC method for semi-supervised medical image segmentation. DuCiSC leverages not only the pixel-level semantic consistency within individual training samples but also the region-level semantic consistency across paired training images. 
    \item We propose dual paradigms to encourage the consistency of region-level semantics by aligning the prototypes of labeled images with unlabeled and fused images, explicitly building up unified semantic representations of them to tackle the distribution mismatch issue. 
    \item We introduce a new self-aware confidence estimation approach to flexibly identify highly-reliable voxels in unlabeled training images, allowing to employ the unlabeled data according to the model’s learning status. 
\end{enumerate}

Extensive experiments on three popular benchmarks (left atrium, pancreas, and ACDC) and a more challenging inferior alveolar nerve dataset reveal the robustness and superiority of our  our DuCiSC method over previous SOTA approaches. 

% The remainder of this paper is organized as follows. 
% Section~\ref{sec:related_work} describes related works; 
% the technical details of our proposed DuCiSC method are introduced in Section~\ref{sec:method}; 
% experimental results on three public benchmarks are reported in Section~\ref{sec:setup} and \ref{sec:results}; 
% and Section~\ref{sec:conclusion} concludes the paper. 

% For consistency with the fused dataset, we introduce the volume fusion strategy to fuse the labeled and unlabeled data which can mitigate the loss of structural information observed in previous methods but maintain the ability to address the distribution mismatch problem. We follow the same protocol to build up the pixel-wise consistency supervised by the fused mask and the semantic-wise prototype consistency between labeled and fused prototypes.

% Additionally, we design a novel uncertainty estimation strategy with a dual dynamic threshold for both foreground and background, which can be dynamically updated and reflect the learning status of the model for the uncertain areas. This approach allows for more accurate identification of reliable regions in the pseudo labels, thereby improving the overall training effectiveness and segmentation performance.

\section{Related Work}
\label{sec:related_work}

\subsection{Semi-supervised Medical Image Segmentation}

% Techniques such as pseudo labeling and consistency regularization are usually adopted for semi-supervised medical image segmentation.
Early efforts on semi-supervised medical image segmentation often rely on the pseudo-labeling technique, where the basic idea is to generate confident pseudo labels on unlabeled training data, by either the model itself~\cite{zhang2022boostmis} or other more robust models~\cite{tarvainen2017mean,cui2019semi}, then these pseudo-labeled training data are further incorporated into the model's learning process~\cite{zhang2022boostmis}. % fan2020inf
% For example, Bai et al.~\cite{bai2017semi} proposed a semi-supervised segmentation method for cardiac images, where pseudo labels were generated by thresholding the model's predicted probabilities. This approach allowed the model to leverage a large amount of unlabeled data, resulting in improved segmentation performance. 
% Similarly, Fan et al.~\cite{fan2020inf} introduced a framework that combined pseudo labeling with uncertainty estimation to enhance the reliability of pseudo labels, further boosting the segmentation accuracy. 
Subsequent studies have shifted towards the consistency regularization technique, due to its outstanding performance and high compatibility with pseudo-labeling. 
This technique focuses on enforcing consistent model outputs under various input or feature perturbations~\cite{yu2019uncertainty,sedai2019uncertainty,xia20203d}, achieved by applying image transformations or injecting random noise. 
In the realm of semi-supervised segmentation, recent works have proposed to enforce the output consistency between separate segmentation networks or heads~\cite{xia2020uncertainty,wu2021semi,wu2022mutual,su2024mutual}. 
For example, UMCT~\cite{xia2020uncertainty} perturbs 3D input volumes into two views and trains two segmentation networks independently on each view, where a co-training strategy is adopted to enforce the multi-view and mutual consistency on unlabeled samples. 
Despite their promising results, these multi-network approaches usually entail increased computation costs for model training.
To deal with this, the MT framework~\cite{tarvainen2017mean} has been gaining popularity. 
This framework self-ensembles the network parameters to create a robust teacher network that is used to generate pseudo labels for unlabeled data~\cite{cui2019semi}.
Based on MT, more advanced consistency-based methods are developed for reaching various objectives, e.g., incorporating self-supervised contrastive regularization~\cite{you2022simcvd}, improving uncertain area selection~\cite{yu2019uncertainty,xu2023ambiguity}, and alleviating labeled-unlabeled distribution mismatch~\cite{bai2023bidirectional} with a bi-directional copy-paste (BCP) strategy. 
Similar to~\cite{bai2023bidirectional}, our DuCiSC also targets overcoming the distribution mismatch problem, 
but differently, we present a more effective approach that directly minimizes the labeled-unlabeled cross-image discrepancy of prototypes representing the region-level semantics and smoothly fuses the labeled and unlabeled images to preserve critical anatomical structures that serve as essential cues in segmenting challenging organs, such as the inferior alveolar nerve.

\subsection{Exploring Cross-image Semantics in Medical Imaging}

Cross-image semantic consistency aims to ensure a unified interpretation or representation of similar structures across different images, which has been proven effective in enhancing the accuracy and reliability of various medical image analysis tasks~\cite{chen2022multi,wu2023cross,wang2024cross,wu2024cephalometric}. 
In organ segmentation, a model trained with cross-image semantic consistency constraints is more likely to precisely segment the anatomical structure despite varying lesion appearances and imaging conditions~\cite{wu2023cross}. 
Meanwhile, maintaining cross-image semantic consistency can enhance the robustness of cephalometric landmark detection across patients of varying age groups~\cite{wu2024cephalometric}. 
To enhance breast cancer detection, semantic consistency across different mammogram views is enforced using multi-view learning techniques~\cite{chen2022multi}. 
In semi-supervised segmentation, the most related works to ours are SCP-Net~\cite{zhang2023self} and CPCL~\cite{xu2022all}.
SCP-Net incorporates both intra-sample and cross-sample consistency within a training mini-batch by leveraging prototypes. 
Specifically, it generates prototypes from both the same sample (intra-sample) and other samples (cross-sample) to create pseudo-label supervisions, which are then employed to optimize the pixel-wise probability predictions for unlabeled training samples. 
Notably, our approach differs from SCP-Net, as we utilize prototypes to enforce region-level semantic alignment between labeled and unlabeled training images, rather than relying on pixel-wise supervision. Furthermore, our training strategy adopts a more comprehensive paradigm, extending beyond the constraints of training mini-batches. 
Similarly, CPCL also introduces prototypes to produce pixel-wise supervision for unlabeled training samples.
To be specific, the similarity between the unlabeled feature maps and labeled prototypes is computed as the pixel-wise guidance for the supervision of the unlabeled sample again. Hence, this strategy is also distinct from our approach that aligns the region-level semantics through prototypes.

\subsection{Confidence Estimation for Pseudo Labeling}

Confidence estimation plays a critical role in applications based on pseudo-labeling techniques, e.g., semi-supervised learning, weakly-supervised learning, and noisy label learning. 
It helps determine the reliability of pseudo labels, 
and in semi-supervised learning, typically requires appropriate thresholds to select confident unlabeled samples used for model training. 
For instance, UDA~\cite{xie2020unsupervised} and FixMatch~\cite{sohn2020fixmatch} employ a fixed, high threshold (e.g., 0.95) across all classes to identify highly-confident training samples. 
However, these approaches have a low data utilization at the early training stage and also overlook the varying learning difficulties between classes. 
This issue has been partly handled by Dash~\cite{xu2021dash} and AdaMatch~\cite{berthelot2021adamatch} that gradually increase the threshold with an ad-hoc scheduler, as the training progresses. 
While enhancing the data utilization, their threshold adjustments are pre-defined with sophisticated hyper-parameters, failing to accurately reflect the model's actual learning status. 
Other approaches for achieving confidence estimation rely on either entropy~\cite{yu2019uncertainty,zhang2023self}, which still presents challenges in determining the appropriate threshold, or the harmonious output of two sub-networks~\cite{su2024mutual}, which can be sensitive to the training data. 
In this paper, we introduce an approach to adjust the confidence threshold in a self-adaptive manner according to the model’s learning status.

\section{Methodology}
\label{sec:method}

\begin{figure*}[ht!]
    \centering
    \includegraphics[width=0.85\linewidth]{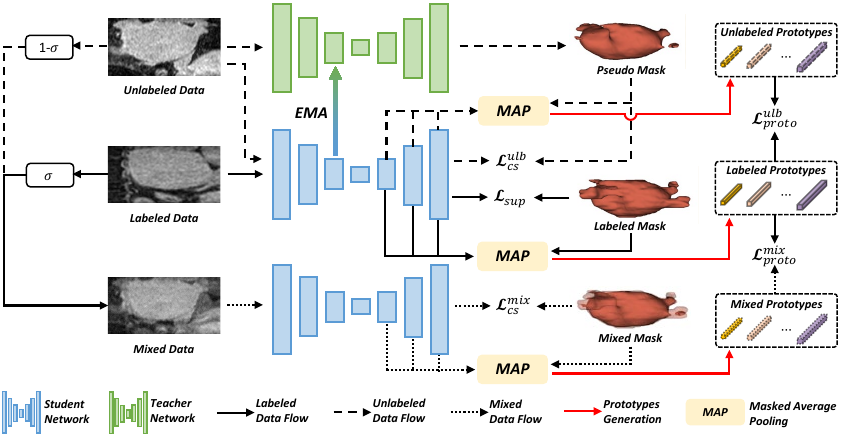}
    \caption{An overview of our DuCiSC method based on MT strategy for semi-supervised medical image segmentation. DuCiSC leverages dual paradigms to enforce the cross-image semantic consistency that is established through region-level prototypes using paired image data: prototype consistency between labeled and unlabeled data ($\mathcal{L}_{proto}^{ulb}$) and prototype consistency between labeled and mixed data ($\mathcal{L}_{proto}^{mix}$). }
    \label{fig:pipeline}
\end{figure*}

We introduce our DuCiSC method for semi-supervised medical image segmentation, 
with the overall framework depicted in Fig.~\ref{fig:pipeline}.
In Section~\ref{sec:preliminary} we first describe our considered problem scenario and provide a preliminary review about the MT strategy, on which our method is based. 
In Section~\ref{sec:dual}, we then propose dual effective paradigms to complementarily enforce the consistency of region-level semantics across different training image pairs, through the representation of prototypes. 
To accurately identify reliable pseudo labels for the model's training, we further present a self-aware confidence estimation approach used to enhance the voxel-wise semantic consistency, which will be elaborated in Section~\ref{sec:dynamic_threshold}.

\subsection{Problem Scenario and MT Preliminary}
\label{sec:preliminary}
For the semi-supervised segmentation task, there is a small amount of labeled data denoted as $\mathcal{D}_{L} = \left\{ \mathbf{x}_i, \mathbf{y}_i \right\}_{i=1}^{|\mathcal{D}_{L}|}$, where $\mathbf{x}_i \in  \mathcal{X} \subset \mathbb{R}^{H \times W \times D}$ represents the 3D input image with size $H \times W \times D$ and $\mathbf{y}_i \in \mathcal{Y} \subset \{0, 1\}^{H \times W \times D \times C}$ denotes the corresponding one-hot ground-truth segmentation map, with $C$ semantic classes to be segmented. 
We also have a large amount of unlabeled data available in $\mathcal{D}_{U} = \left\{ \mathbf{x}_j \right\}_{j=1}^{|\mathcal{D}_{U}|}$, where $|\mathcal{D}_{U}| \gg |\mathcal{D}_{L}|$.
Samples from both $\mathcal{D}_{L}$ and $\mathcal{D}_{U}$ are leveraged to train a segmentation model (e.g., V-Net~\cite{milletari2016v}) $f_{\theta}:\mathcal{X} \to \mathcal{Y}$, with $\theta$ denoting the model's parameter. 
% Typically, the segmentation model can be decomposed as an encoder $h_{\theta_h}:\mathcal{X} \to \mathcal{F}$ to extract multi-level feature representations $\mathbf{F}_E = \{\mathbf{F}^1_E,..., \mathbf{F}^S_E\} = h_{\theta_h}(\mathbf{x})$, where $\mathbf{F}^s_E \in \mathbb{R}^{H^s \times W^s \times D^s \times Z^s }$ denoting the encoded feature maps (with $Z^s$ feature channels) at scale $s$, as well as a decoder $g_{\theta_g}:\mathcal{F} \to \Delta$ to map the concatenated \chong{} feature representations $\mathbf{F}$ upsampled and to the softmax segmentation predictions $\hat{\mathbf{y}}= g_{\theta_g}(\mathbf{F}) \in \Delta \subset {[0,1]}^{H \times W \times D \times C}$. 
Typically, the segmentation model has an encoder-decoder architecture, where the encoder progressively compresses the spatial dimensions and the decoder gradually restores the spatial resolutions, producing multi-level feature representations $\mathbf{F} = \{\mathbf{F}^1,..., \mathbf{F}^S\}$, 
with $\mathbf{F}^s \in \mathbb{R}^{H^s \times W^s \times D^s \times Z^s }$ denoting the decoded feature maps (with $Z^s$ feature channels) at scale $s$. 
Usually, the feature at the highest level is employed to generate the probabilistic segmentation predictions 
by applying the softmax function $\delta(\cdot)$ over $C$ classes: $\hat{\mathbf{y}} = \delta({\mathbf{F}^S)} \in {[0,1]}^{H \times W \times D \times C}$. 

% These feature representations are further processed by up-sampling, skip-connection to produce the probabilistic segmentation predictions $\hat{\mathbf{y}} \in {[0,1]}^{H \times W \times D \times C}$, where to produce the probability map. 

% The encoder gradually reduces the spatial dimension of the pooling layer.
% The decoder gradually repairs the details and spatial dimensions of the object. 
% There is usually a skip connection between the encoder and the decoder, which helps the decoder to accurately reconstruct the target.

% For convenience, we use binary segmentation as an example, but this approach can be easily extended to multi-class segmentation tasks.

The MT framework is a classical and popular framework used for semi-supervised learning tasks~\cite{tarvainen2017mean,cui2019semi}, where its training objective can be formulated as:
\begin{equation}
\begin{aligned}
    \min_{\theta} & \sum_{i=1}^{|\mathcal{D}_{L}|} \mathcal{L}_{sup}(f(\mathbf{x}_{i}; \theta),\mathbf{y} _{i}) + \sum_{j=1}^{|\mathcal{D}_{L}|+|\mathcal{D}_{U}|} \mathcal{L}_{cs}(f(\mathbf{x}_{j}; \theta), \bar{\mathbf{y}}_j), 
\end{aligned}
\label{eq:MT}
\end{equation}
% \chong{can we remove $\lambda$ in the above equation, because we never use it in the following paper.}\han{I think it's ok.}
where $\mathcal{L}_{sup}(\cdot)$ is the voxel-wise supervised loss (typically defined as a combination of Dice and cross-entropy losses~\cite{yu2019uncertainty,bai2023bidirectional,xu2023ambiguity}), to train the current student model $f_{\theta}$ on labeled training samples from $\mathcal{D}_{L}$. 
$\mathcal{L}_{cs}(\cdot)$ is the voxel-wise consistency loss to supervise the student's predictions on unlabeled and labeled training samples in $\mathcal{D}_{U} \cup \mathcal{D}_{L}$, which are pseudo-labeled by another teacher model $f_{\bar{\theta}}$ with: $\bar{\mathbf{y}}_j = \argmax_c f_{\bar{\theta}}(\mathbf{x}_{j})$. 
% where $\eta(\cdot)$ applies perturbation noise. 
The teacher model usually has the same network structure with the current student model. 
During training, the student model is optimized according to Eq.~(\ref{eq:MT}), whereas the teacher model is updated by the exponential moving average (EMA) of the parameter of the student model, i.e., $\bar{\theta}_t = \alpha \bar{\theta}_{t-1} + (1 - \alpha) \theta_t$, 
where $t$ denotes the training iteration step and $\alpha$ represents the smoothing coefficient of EMA.

According to Eq.~(\ref{eq:MT}), the current MT-based approaches in semi-supervised medical image segmentation are limited in the following aspects.  
On the one hand, their training only considers intra-image voxel-wise semantics in both terms of Eq.~(\ref{eq:MT}), neglecting a more comprehensive understanding of the region-level semantics.
Also, they exploit the labeled and unlabeled training data in a separate learning scheme~\cite{bai2023bidirectional}, 
adversely causing the labeled-unlabeled distribution mismatch. 
To tackle these, we propose to make full use of the region-level semantics by enforcing semantic consistency across dual pairs of training images, elaborated in Section~\ref{sec:dual}. 
On the other hand, for the voxel-wise consistency regularization defined in the second term of Eq.~(\ref{eq:MT}), early studies~\cite{perone2018deep,cui2019semi} have no mechanism to exclude unreliable pseudo labels in $\bar{\mathbf{y}}_j$ for the student's learning, leading to severe confirmation bias~\cite{arazo2020pseudo} and inferior generalization ability. 
Recent works~\cite{yu2019uncertainty,xu2023ambiguity} present remedies with predefined constant or entropy-based confidence thresholds, which are not effective in capturing the training dynamics of the class-wise confidence. 
In this paper, we address this issue by proposing a new self-aware confidence estimation strategy that flexibly selects sufficiently reliable pseudo labels, explained in Section~\ref{sec:dynamic_threshold}. 

% \chong{intra-image pixel-wise semantics, cross-image region-level semantics}

\subsection{Dual Cross-image Semantic Consistency}
\label{sec:dual}

One significant challenge in semi-supervised medical image segmentation is the undesirable discrepancy in the extracted features between labeled and unlabeled training samples, 
which is evidenced by the empirical distribution mismatch between the two sets of training data~\cite{bai2023bidirectional}. 
To overcome this issue, in this work we present dual paradigms to effectively enforce the region-level semantic consistency between:
1) labeled and unlabeled training images (Section~\ref{sec:lab&unlabl}); 
and 2) labeled and synthetic training images (Section~\ref{sec:lab&fuse}).

\subsubsection{Semantic Consistency Across Labeled and Unlabeled Images}
\label{sec:lab&unlabl}

% \chong{please provide a figure about how to generate the prototypes, such as masked average pooling.}

In medical image segmentation, there is a common observation that pixels or voxels of the same class are prone to share similar feature representations~\cite{lee2022voxel,wang2023learning,he2023segmentation} since they are supervised to represent the same object (e.g., organ). 
Leveraging such observation, we can establish an effective representation of the region-level semantics by extracting the average class-wise image features. 
To be concrete, we employ the masked average pooling~\cite{du2022weakly} technique to produce prototypes that serve as the region-level semantics:
\begin{equation}
    \mathbf{p}_i^{c, s} =  \frac{ \sum_{(h, w, d)}^{(H^s, W^s, D^s)} \mathbf{y}^c_i(h, w, d) \cdot \mathbf{F}^s_i(h, w, d)}{\sum_{(h, w, d)}^{(H^s, W^s, D^s)} \mathbf{y}^c_i(h, w, d)},
\label{fully_prototypes}
\end{equation}
% where $(h, w, d)$ denotes the spatial indexes, $\mathbf{F}^s_i$ is the feature maps at scale $s$ computed by the \chong{student} encoder from a labeled training image $\mathbf{x}_i$, and $\mathbf{y}^c_i$ is the corresponding ground-truth segmentation map of class $c$.
where $(h, w, d)$ denotes the spatial indexes, $\mathbf{F}^s_i$ is the feature maps at scale $s$ extracted from a labeled training image $\mathbf{x}_i$, and $\mathbf{y}^c_i$ is the corresponding ground-truth segmentation map of class $c$.
The generated prototypes $\mathbf{p}_i^{c, s} \in \mathbb{R}^{1 \times 1 \times 1 \times Z^s}$ are capable of holistically capturing class-representative features with common characteristics for a particular class (typically an organ object) in $\mathbf{x}_i$.  
It is worth mentioning that the segmentation map $\mathbf{y}^c_i$ may have a different spatial resolution from the feature maps $\mathbf{F}^s_i$, which potentially necessitates an additional rescaling operation for the segmentation map. 
Analogous to the labeled sample $\mathbf{x}_i$, we can also extract prototypes from an unlabeled training image $\mathbf{x}_j$, 
in which the predicted segmentation maps $\bar{\mathbf{y}}_j$, pseudo-labeled by the teacher model, is harnessed as follows: 
\begin{equation}
    \mathbf{p}_j^{c,s} = \frac{ \sum_{(h, w, d)}^{(H^s, W^s, D^s)} \bar{\mathbf{y}}^c_j(h, w, d) \cdot \mathbf{F}^s_j(h, w, d)}{\sum_{(h, w, d)}^{(H^s, W^s, D^s)} \bar{\mathbf{y}}^c_j(h, w, d)},
\label{unlabeled_prototypes}
\end{equation}
% where $\mathbf{F}^s_j$ denotes the latent feature maps at scale $s$ computed by the \chong{student} encoder from an unlabeled sample $\mathbf{x}_j$. 
where $\mathbf{F}^s_j$ denotes the feature maps at scale $s$ computed from an unlabeled training sample $\mathbf{x}_j$. 

Based on Eq.~(\ref{fully_prototypes}) and Eq.~(\ref{unlabeled_prototypes}), 
% the consistency regularization should be applied to not only the pixel-wise semantics for individual images, 
we propose to enforce consistency of the region-level semantics extracted from the labeled and unlabeled image pair,
which is accomplished by aligning their class-wise prototypes, as follows: 
\begin{equation}
\begin{aligned}
\mathcal{L}_{proto}^{ulb} = \frac{1}{C} \sum_{s=1}^S \sum_{c=1}^C || \mathbf{p}_i^{c,s} - \mathbf{p}_j^{c,s}||^2_2.
\end{aligned}
\label{proto_unlabled}
\end{equation}
Relying on Eq.~(\ref{proto_unlabled}), our segmentation model can not only comprehensively understand the image semantics at a higher image-region level but also explicitly build the consistency relationship between labeled and unlabeled training images.

% \subsection{labeled-unlabeled matching with Volume Fusion}
\subsubsection{Semantic Consistency Across Labeled and Fused Images}
\label{sec:lab&fuse}

To deal with the feature discrepancy issue, our proposal in Section~\ref{sec:lab&unlabl} is obtaining consistent semantics between labeled and unlabeled images, which is implemented by the region-based prototypes solely in the feature space. 
The recent BCP approach~\cite{bai2023bidirectional} suggests a copy-paste strategy (e.g., CutMix~\cite{yun2019cutmix}) to bidirectionally fuse labeled images with unlabeled images to create new mixed training samples on which the model predictions are supervised by the mixed signals accordingly. 
This is a straightforward and useful idea to mitigate the unlabeled-unlabeled distribution gap, 
as training on the mixed samples helps the model learn common semantics between the labeled and unlabeled data. 
However, the BCP approach requires cropping the image patch with an appropriate size that is hard to determine, 
the copy-paste strategy could also potentially destroy the underlying anatomical structures, 
which are critical cues in segmenting challenging organs, e.g., the inferior alveolar nerve that has tube-like shape and small size. 
% \chong{predefined fixed crop size, no dynamic fusion}
% Existing works~\cite{sohn2020fixmatch,wang2023freematch} have extensively demonstrated that creating synthetic training data with augmentation techniques, e.g., Mixup~\cite{zhang2018mixup} and CutMix~\cite{yun2019cutmix}, can greatly benefit the performance of semi-supervised tasks, due to the diversifying of unlabeled training data. 
To this end, we introduce a natural way to seamlessly fuse two images, 
relying on the Mixup~\cite{zhang2018mixup} that linearly interpolates between a labeled image $\mathbf{x}_i$ and an unlabeled image $\mathbf{x}_j$:
\begin{equation}
\begin{aligned}
    \mathbf{x}_k &= \sigma \cdot \mathbf{x}_i + (1-\sigma) \cdot \mathbf{x}_j, \\
    \mathbf{y}_k &= \sigma \cdot \mathbf{y}_i + (1-\sigma) \cdot \bar{\mathbf{y}}_j,
\end{aligned}
\label{eq:mixup}
\end{equation}
% \chong{please experiment with combining two unlabeld images...}\han{Ongoing. Done, no big difference.}
% Mixup [229] performs linear interpolations of two inputs and their corresponding labels. 
% This simple technique imposes consistency regularization to guide the learning of a mapping between the interpolated input and interpolated output to learn from unlabeled data.
where $\sigma$ is a combination ratio sampled from a uniform distribution, i.e, $\sigma \sim \mathcal{U}(0.25, 0.75)$ aiming at offering a sufficiently strong mixture effect, 
and $\mathbf{y}_k$ denotes the fused ``soft'' pseudo label utilized for supervising the predictions on mixed training image $\mathbf{x}_k$, 
described in the next Section~\ref{sec:dynamic_threshold}. 
The mixed training image $\mathbf{x}_k$ is then incorporated into the student network to produce the feature representations $\mathbf{F}^s_k$ at different feature scales $s$, 
on which we can readily extract the region-level semantics (i.e, prototypes), with: 
\begin{equation}
\begin{aligned}
    \mathbf{p}_k^{c,s} = \frac{\sum_{(h, w, d)}^{(H^s, W^s, D^s)} \mathbf{y}^c_k(h, w, d) \cdot \mathbf{F}^s_k(h, w, d)}{\sum_{(h, w, d)}^{(H^s, W^s, D^s)} \mathbf{y}^c_k(h, w, d)}.
\end{aligned}
\end{equation}
Eventually, we minimize the following region-level semantic consistency loss to improve the class-wise prototype alignment between the labeled and mixed training images: 
\begin{equation}
    \mathcal{L}_{proto}^{mix} = \frac{1}{C} \sum_{s=1}^S \sum_{c=1}^C || \mathbf{p}_i^{c,s} - \mathbf{p}_k^{c,s}||^2_2.
    \label{eq:prototype_fused}
\end{equation}

It is noteworthy that our primary goal of utilizing fused images is establishing the cross-image semantic consistency in Eq.~(\ref{eq:prototype_fused}) to tackle the distribution mismatch, 
which differs from previous methods that apply data augmentations on unlabeled images to achieve voxel-wise semantic consistency with pseudo-labeling supervision (as the second term of Eq.~(\ref{eq:MT})). 
Also, BCP~\cite{bai2023bidirectional} considers only voxel-wise semantic consistency 
but overlooks the region-level semantic consistency in Eq.~(\ref{eq:prototype_fused}),  
despite creating and using mixed training samples.

Relying on our dual paradigms of the cross-image semantic consistency,
our segmentation model is likely to build a broad spectrum of consistent relationships efficiently, 
due to the utilization of paired samples, whose amount is approximate to the square of the number of labeled training images and the number of unlabeled training images.
In addition, the paired samples are further increased significantly, due to the random mixture of labeled and unlabeled training images.

% CutMix [42] is simple yet strong data processing method, also dubbed as Copy-Paste (CP), which has the potential to encourage unlabeled data to learn common semantics from the labeled data, since pixels in the same map share semantics to be closer [29]. 
% In semi-supervised learning, forcing consistency between weak-strong augmentation pair of unlabeled data is widely used [11, 14, 32, 47], and CP is usually used as a strong augmentation. 
% But existing CP methods only consider CP cross unlabeled data [8, 10, 14], or simply copy crops from labeled data as foreground and paste to another data [6, 9].
% They neglect to design a consistent learning strategy for labeled and unlabeled data, which hampers its usage on reducing the distribution gap. 

% Meanwhile, CP tries to enhance the generalization of networks by increasing unlabeled data diversity, but a high performance is hard to achieve since CutMixed image is only supervised by low-precision pseudo-labels. 
% It’s intuitive to use more accurate supervision to help networks segment degraded region cut by CP.

% \subsection{Dual-Dynamic Threshold-based Uncertainty Estimation}
% \subsection{Adaptive Uncertainty Estimation}
% \subsection{Self-aware Pseudo Labeling}

\subsection{Intra-image Semantic Consistency by Self-aware Pseudo Labeling}
\label{sec:dynamic_threshold}

In addition to the cross-image region-level semantic consistency, 
our DuCiSC method also enhances the intra-image voxel-wise semantic consistency for the unlabeled and fused images, based on the pseudo-labeling supervision in the second term of Eq.~(\ref{eq:MT}). 
Ideally, this demands effective mechanisms to ensure that only sufficiently reliable pseudo-labeled voxels are involved in the student's learning process~\cite{arazo2020pseudo}.
Previous approaches to select these reliable voxels typically depend on confidence thresholds, 
which are pre-defined for all classes~\cite{mahmood2024splal} or evaluated using model's prediction entropy computed on individual training samples~\cite{yu2019uncertainty,xu2023ambiguity}, 
thereby failing to consider the actual learning status of different classes. 
In this paper, we address this by presenting a new self-aware confidence threshold estimation approach that makes full use of the training dynamics of unlabeled data.
Concretely, for an unlabeled training image $\mathbf{x}_j$, 
% we select the voxels predicted as class $c$ by the student model and estimate their average probability $P^{c}_{\text{avg}}$, 
we first compute the student network's average probability $P^{c}_{\text{avg}}$ on voxels that are selected by the teacher's segmentation mask.  
Subsequently, the probability $P^{c}_{\text{avg}}$ from individual samples is accumulated to update the class-wise confidence threshold $T^{c}_t$ with an EMA fashion at each training iteration step $t$: 
\begin{equation}
\begin{aligned}
T_t^{c}= 
\left\{ 
    \begin{array}{ll}
        \nicefrac{1}{C},   \qquad \ \ \ \ \ \ \ \ \ \ \ \ \ \ \ \ \ \ \,         {\rm if} \ t=0,  \\
        \beta T^{c}_{t-1} + (1 - \beta) P^{c}_{\text{avg}}, \ \ \                {\rm otherwise}, \\
    \end{array}
\right.
\end{aligned}
\label{eq:thresholds}
\end{equation}
where $\beta$ is a smoothing factor. We leverage the EMA technique, as it incorporates a great amount of historical information as the training evolves, ensuring the robustness of the confidence threshold estimation.
$P^{c}_{\text{avg}}$ is calculated as:
\begin{equation}
\begin{aligned}
    P^{c}_{\text{avg}} = \frac{\sum_{(h, w, d)}^{(H, W, D)} \bar{\mathbf{y}}_j^c(h, w, d) \cdot \hat{\mathbf{y}}^c_j(h, w, d)}{\sum_{(h, w, d)}^{(H, W, D)} \bar{\mathbf{y}}_j^c(h, w, d)},
\end{aligned}
\label{eq:avg_prob}
\end{equation}
where $\hat{\mathbf{y}}^c_j$ represents the softmax probability of class $c$ from the student network
and $\bar{\mathbf{y}}^c_j$ denotes the one-hot pseudo label of class $c$ predicted by the teacher network, 
both for the unlabeled training image $\mathbf{x}_j$. 
In Eq.~(\ref{eq:avg_prob}), $\bar{\mathbf{y}}^c_j$ can be viewed as a selector to aggregate the corresponding probabilities in $\hat{\mathbf{y}}^c_j$.
Notably, our confidence estimation approach is entirely self-aware and does not introduce additional parameters.

Leveraging Eq.~(\ref{eq:thresholds}), we can formulate the following intra-image semantic consistency loss $\mathcal{L}_{cs}^{ulb}(\cdot)$ and $\mathcal{L}_{cs}^{mix}(\cdot)$ at the voxel level for unlabeled images and mixed images, respectively, according to the second term in Eq.~(\ref{eq:MT}):
\begin{equation}
\begin{aligned}
\mathcal{L}_{cs}^{ulb} = \ell_{Dice}(f(\mathbf{x}_{j}; \theta), \bar{\mathbf{y}}_j, T) + \ell_{CE}(f(\mathbf{x}_{j}; \theta), \bar{\mathbf{y}}_j, T), \\
\mathcal{L}_{cs}^{mix} = \ell_{Dice}(f(\mathbf{x}_{k}; \theta), \mathbf{y}_k, T) + \ell_{CE}(f(\mathbf{x}_{k}; \theta), \mathbf{y}_k, T),
\end{aligned}
\end{equation}
where ${T} = \{ T_t^1, ..., {T}_t^{C} \}$ denotes our self-aware confidence thresholds for all $C$ classes, used for selecting areas with highly-reliable pseudo labels in $\bar{\mathbf{x}}_j$ and $\mathbf{x}_k$ to contribute to the student's training, 
where $\mathbf{x}_k$ shares the same thresholds with $\mathbf{x}_j$, 
considering that $\mathbf{x}_k$ is derived from the mixture of the labeled image $\mathbf{x}_i$ and unlabeled image $\mathbf{x}_j$. 
$\mathcal{L}_{Dice}(\cdot)$ and $\mathcal{L}_{CE}(\cdot)$ compute the Dice loss and cross-entropy (CE) loss, respectively.

At the beginning of training, our self-aware threshold $T$ is low in order to include more potentially correct voxels for model training. 
As the model become more confident, the threshold is gradually increased to filter out incorrect voxels, thereby minimizing the risk of confirmation bias.

\subsection{Overall Training Objectives}
\label{sec:overallObjectives}

Relying on the cross-image region-level semantic consistency and intra-image voxel-level semantic consistency, the overall training objective of our DuCiSC is defined as:
\begin{equation}
\begin{aligned}
\mathcal{L} = \mathcal{L}_{sup} + \lambda_1 \mathcal{L}_{proto}^{ulb} + \lambda_2 \mathcal{L}_{proto}^{mix} 
+ \lambda_3 \mathcal{L}_{cs}^{ulb} + \lambda_4 \mathcal{L}_{cs}^{mix}
\end{aligned}
\label{eq:loss}
\end{equation}

\section{Experiment Setups}
\label{sec:setup}
\subsection{Datasets}
To evaluate the effectiveness of our method, we conduct extensive experiments on four public datasets including three 3D binary medical image datasets: Left Atrium (LA)~\cite{xiong2021global}, Pancreas~\cite{clark2013cancer}, and Inferior Alveolar Nerve (IAN)~\cite{cipriano2022improving}, and one 2D multi-class medical image datasets: Automatic Cardiac Diagnosis Challenge (ACDC)~\cite{bernard2018deep} dataset.

\subsubsection{LA dataset} LA~\cite{xiong2021global} is a benchmark segmentation dataset from the 2018 Atrial Segmentation Challenge, which comprises 100 3D left atrium scans acquired by cardiac magnetic resonance imaging (MRI) with an isotropic resolution of $0.625\times0.625\times0.625$ $mm^3$. By strictly following the data split protocol and pre-processing procedure in previous approaches~\cite{yu2019uncertainty,li2020shape,xiong2021global}, we train our model with 10\% and 20\% annotations from the same 80 scans and evaluate it on the remaining 20 scans. Our model employs 112 × 112 × 80 image patches for both training and testing, and a stride of 18 × 18 × 4 is used for sliding-window inference. 
% \chong{what is the labeled and unlabeled ratio ?}\han{the percentage of used labeled data like 10\% or 20\%}

\subsubsection{NIH-Pancreas dataset} NIH-Pancreas~\cite{clark2013cancer} has 82 3D abdominal contrast-enhanced CT scans, from 
% which are collected from 53 male and 27 female subjects 
the National Institutes of Health (NIH). 
These scans have the same resolution of 512 × 512 with varying thicknesses from 1.5 to 2.5 $mm$. 
Following the same data split in~\cite{luo2021semi,su2024mutual}, we adopt 62 scans for training and report performances on the remaining 20 scans. We randomly extract 3D patches of size 96 × 96 × 96 as the model input in training and use a sliding-window strategy with a stride of 16 × 16 × 16 for testing. We employ 10\% and 20\% of the training data as labeled training samples while the rest of the training set is regarded as unlabeled samples.

\subsubsection{Inferior Alveolar Nerve dataset} This dataset originates from~\cite{cipriano2022improving} and the recent MICCAI 2023 Challenge\footnote{https://toothfairy.grand-challenge.org/}, which totally has 153 3D CBCT scans with voxel-wise annotations of the inferior alveolar nerve. These scans have the same thickness of 0.3 $mm$ but different spatial resolutions ranging from 148 × 265 × 312 to 178 × 423 × 463. 
In this work, we randomly chose 110 scans and split them into 90 training and 20 testing samples. 
The patch size is 160 × 128 × 112 and a stride of 16 × 16 × 16 is used for sliding-window inference. 
We experiment with 10\% and 20\% training data as labeled images and the remaining as unlabeled images.
It is important to note that the inferior alveolar nerve presents a more challenging segmentation task due to its tube-like shape and small size compared with the left atrium and pancreas. 

\subsubsection{ACDC dataset} 
ACDC~\cite{bernard2018deep} is often used as a 2D benchmark with four classes (i.e., background, right ventricle, left ventricle, and myocardium), which contains 100 cardiac MR volumes. As in prior studies, all images are treated as 2D slices and resized to 256 × 256 with normalization to [0, 1].

\input{results/LA_new}

\subsection{Evaluation Metrics} 
\label{sec:metrics}
To evaluate the segmentation performance, we use the following metrics~\cite{yu2019uncertainty,wu2022exploring,wu2022mutual,su2024mutual}: Dice, Jaccard, average surface distance (ASD), and 95\% Hausdorff Distance (95HD)
where Dice and Jaccard are measured in percentage, while ASD and 95HD are measured in voxels. 

To evaluate the effect of labeled-unlabeled distribution matching, 
we first follow the kernel density estimation (KDE) technique~\cite{bai2023bidirectional}, which is performed at the highest feature level $\mathbf{F}^{S}$, to visualize the feature distribution differences between labeled and unlabeled training samples.
Additionally, we also propose a new quantitative measure to assess the whole-set labeled-unlabeled semantic matching matrix $\mathbf{M} \in \mathbb{R}^{|\mathcal{D}_{L}| \times |\mathcal{D}_{U}| \times C}$, with each element calculated by: 
\begin{equation}
    \mathbf{M}^{c}(i, j) = \exp^{-|| \mathbf{p}_i^{c,S} - \mathbf{p}_j^{c,S}||^2_2},
    \label{eq:prototype_matrix}
\end{equation}
where $\mathbf{p}_i^{c,S}$ and $\mathbf{p}_j^{c,S}$ represent the prototypes (at the highest feature level $S$) computed from a labeled and unlabeled training sample, respectively.  
Obviously, a larger element in $\mathbf{M}$ means better semantic matching. 
Relying on $\mathbf{M}$, we further compute the whole-set labeled-unlabeled semantic matching score $Q = \frac{1}{C\times |\mathcal{D}_{L}| \times |\mathcal{D}_{U}|}\sum_c^C \sum_{i}^{|\mathcal{D}_{L}|} \sum_{j}^{ |\mathcal{D}_{U}|} \mathbf{M}^{c}(i, j)$.

\subsection{Implementation Details}
% Following \cite{yu2019uncertainty,su2024mutual},
We employ V-Net~\cite{milletari2016v} as the model backbone to construct our DuCiSC in all the experiments. 
To enhance convergence, we harness deep supervision on each stage of the decoder by appending an extra $1 \times1 \times 1$ convolution layer.
For computing prototypes from multi-level features, we utilize a total of $S$ = 4 feature levels. 
In each training mini-batch, we use 2 labeled images and 2 unlabeled images. 
Our DuCiSC model is trained using a stochastic gradient descent (SGD) optimizer with a momentum of 0.9, initial learning rate of $1.0 \times 10^{-2}$, weight decay of $5.0 \times 10^{-4}$, and maximum training iteration number of $1.5 \times 10^{4}$. 
To achieve a fair comparison, we follow the same online data augmentation in~\cite{yu2019uncertainty,su2024mutual}. 
In Eq.~(\ref{eq:loss}), we have $\lambda_1 = \lambda_2 = \lambda_4 = 0.1$ and $\lambda_3 = 0.3$ for all datasets.
% with an ablation study given in Fig.~\ref{}. 
All experiments are implemented in Pytorch on an NVIDIA Tesla A100 (40GB) GPU.

\section{Experiment Results}
\label{sec:results}
\subsection{Comparison with SOTA Approaches}

We compare the proposed DuCiSC with previous SOTA semi-supervised segmentation methods, including UA-MT~\cite{yu2019uncertainty}, DTC~\cite{luo2021semi}, MCNet~\cite{wu2021semi}, MCNet+~\cite{wu2022mutual}, SimCVD~\cite{you2022simcvd}, CAML~\cite{gao2023correlation}, BCP~\cite{bai2023bidirectional}, EIC~\cite{huang2024exploring}, MLRPL~\cite{su2024mutual} and TraCoCo~\cite{liu2024translation}. 
We have also included recent dual-teacher methods: PS-MT~\cite{liu2022perturbed} and AD-MT~\cite{zhao2024alternate}, and FixMatch-based methods: FixMatch~\cite{sohn2020fixmatch}, UniMatch~\cite{yang2023revisiting} and DistillMatch~\cite{wang2024distillmatch}.
We also compare with the fully-supervised setting, which only uses the labeled data to train the original V-Net~\cite{milletari2016v} (3D scenario) and U-Net~\cite{ronneberger2015u} (2D scenario).

\begin{figure}[t!]
    \centering
    \includegraphics[width=1.0\linewidth]{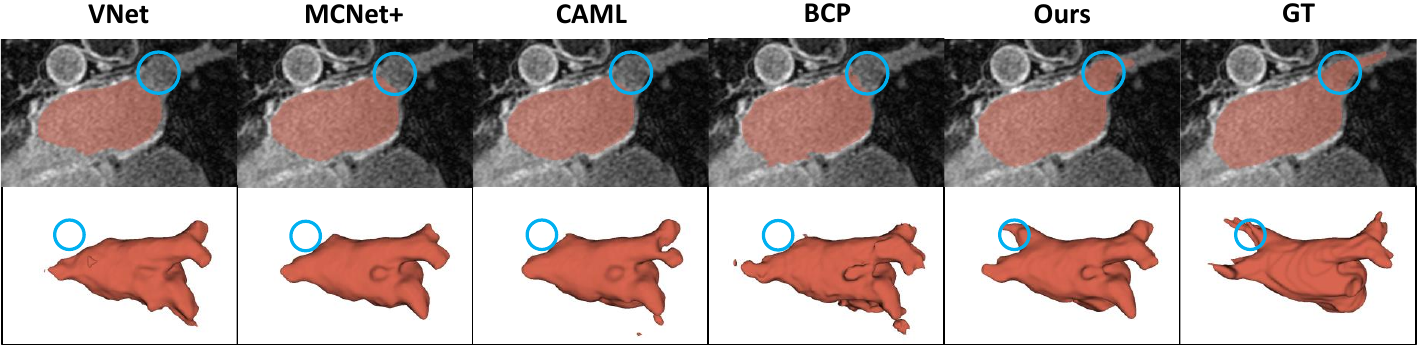}
    \caption{Visual results of LA segmentation between our method and other SOTA approaches, trained on 10\% labeled data. }
    \label{fig:LA_8}
\end{figure}

\subsubsection{Performance on LA dataset} 

Our DuCiSC method is first evaluated on the LA dataset, with results given in Table~\ref{tab:la_results}. 
As evident, DuCiSC exhibits the best segmentation results in all evaluation metrics (Dice, Jaccard, 95HD, and ASD) and experimental protocols (both 10\% and 20\% labeled data). 
Specifically, DuCiSC achieves a Dice score of 91.81\% with a Jaccard score of 84.93\% using 10\% labeled data, and a Dice score of 92.11\% with a Jaccard score of 85.42\% using 20\% labeled data, 
surpassing other advanced approaches, such as MLRPL~\cite{su2024mutual}, EIC~\cite{huang2024exploring}, and BCP~\cite{bai2023bidirectional}, by large margins.  
In Fig.~\ref{fig:LA_8}, we present a typical visual segmentation example from V-Net, MCNet+, CAML, BCP, and our DuCiSC method. 
These compared methods are prone to produce anatomically implausible predictions with erroneous segmentations (indicated by the blue circle). 
In contrast, our DuCiSC yields predictions that align more closely with the ground-truth (GT) segmentation, offering superior anatomical plausibility.

\begin{figure}[t!]
    \centering
    \includegraphics[width=1.0\linewidth]{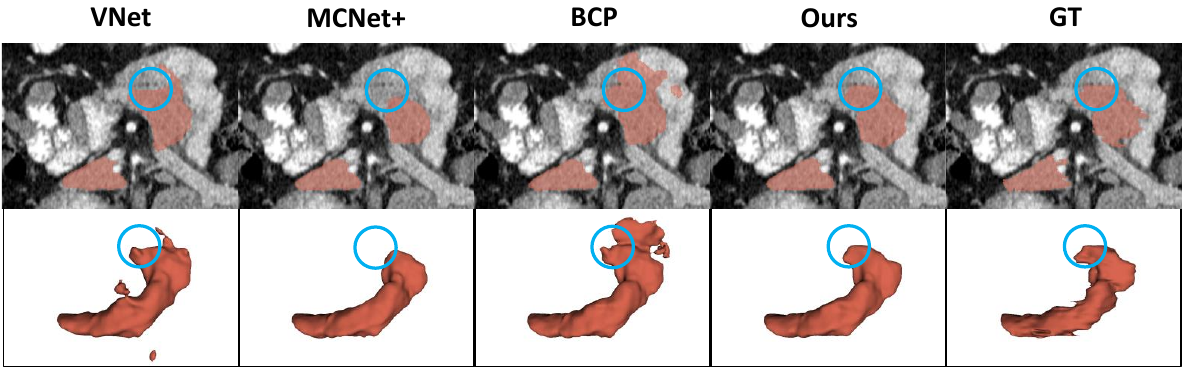}
    \caption{Segmentation visualization of pancreas from our method and other SOTA approaches, trained 20\% labeled data. }
    \label{fig:Pancreas_12}
\end{figure}
\subsubsection{Performance on Pancreas dataset} 

\input{results/Pancreas_new}

Table \ref{tab:pan_results} illustrates the quantitative segmentation results of our DuCiSC and other competing methods, 
where we note DuCiSC obtains substantial performance gains when using 10\% labeled and 90\% training data, 
demonstrating its advantage in situations where labeled data are extremely limited.
Substantial improvements can also be observed under the setting of 20\% labeled with 80\% unlabeled data. 
We display the visual segmentation results in Fig. \ref{fig:Pancreas_12}, 
showing that DuCiSC segments the whole pancreas region more accurately than other compared methods.

\input{results/Nerves_results}

\input{results/ACDC}

\subsubsection{Performance on IAN dataset}
\begin{figure*}[h!]
    \centering
    \includegraphics[width=1.0\linewidth]{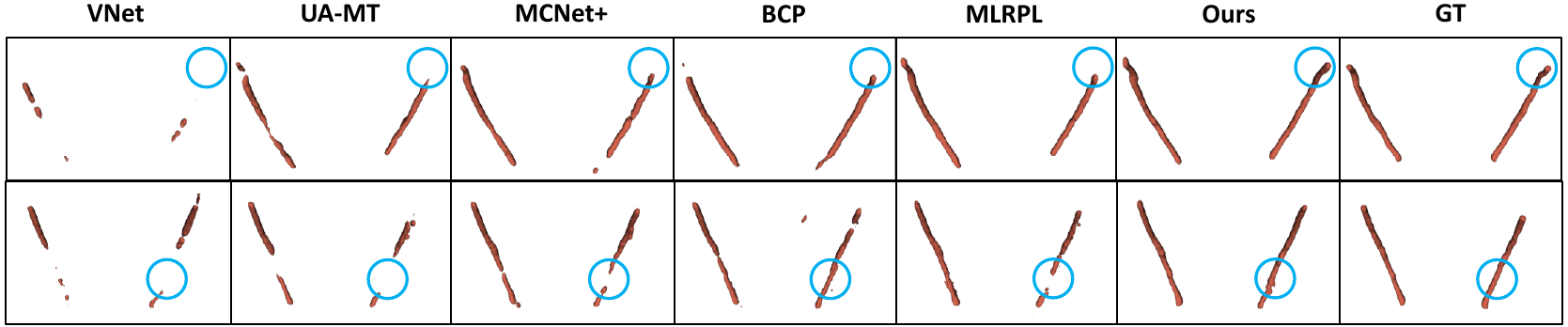}
    \caption{Typical visual segmentation results of the inferior alveolar nerve from different approaches, which are trained using 10\% labeled data in IAN dataset. Each row shows a different case. }
    \label{fig:Nerve_9}
\end{figure*}

Segmenting the inferior alveolar nerve is a newly proposed challenging topic to evaluate the model's generalizability and robustness to organs with tube-like shape and small size. 
In this experiment, we implement four competing approaches that are top-performing on LA and Pancreas tasks:
UA-MT~\cite{yu2019uncertainty}, MCNet+~\cite{wu2022mutual}, BCP~\cite{bai2023bidirectional} and MLRPL~\cite{huang2024exploring}. 
The segmentation results are reported in Table~\ref{tab:nerves_results}. 
It is noticed that in the two semi-supervised settings (10\% labeled, 90\% unlabeled) and (20\% labeled, 80\% unlabeled),
our DuCiSC consistently outperforms other competing methods with Dice improvements of 5.69\% and 2.13\%, respectively, with respect to the second best approach, MLRPL~\cite{huang2024exploring}.
Fig.~\ref{fig:Nerve_9} illustrates a visual comparison of all approaches in segmenting two inferior alveolar nerve cases, 
where the segmentation predictions from the V-Net baseline are severely fragmented and all the semi-supervised methods can segment a more complete structure for the inferior alveolar nerve. 
However, these competing methods still suffer from wrong segmentation in certain areas marked by the circles (particularly occurs in the end parts of the nerve), 
whereas our DuCiSC approach demonstrates superior accuracy.

\subsubsection{Performance on ACDC dataset}
To further validate the generalization ability and robustness of our DuCiSC approach on multi-class segmentation tasks, we conducted additional experiments using the ACDC dataset, with results given in Table \ref{tab:ACDC}, 
where DuCiSC obtains consistent performance gains, 
demonstrating its advantage in situations where labeled training data is limited in a multi-class segmentation task.

\subsection{Evaluation of Unlabeled-labeled Distribution Matching}

\begin{figure*}[h!]
    \centering
    \includegraphics[width=1.0\linewidth]{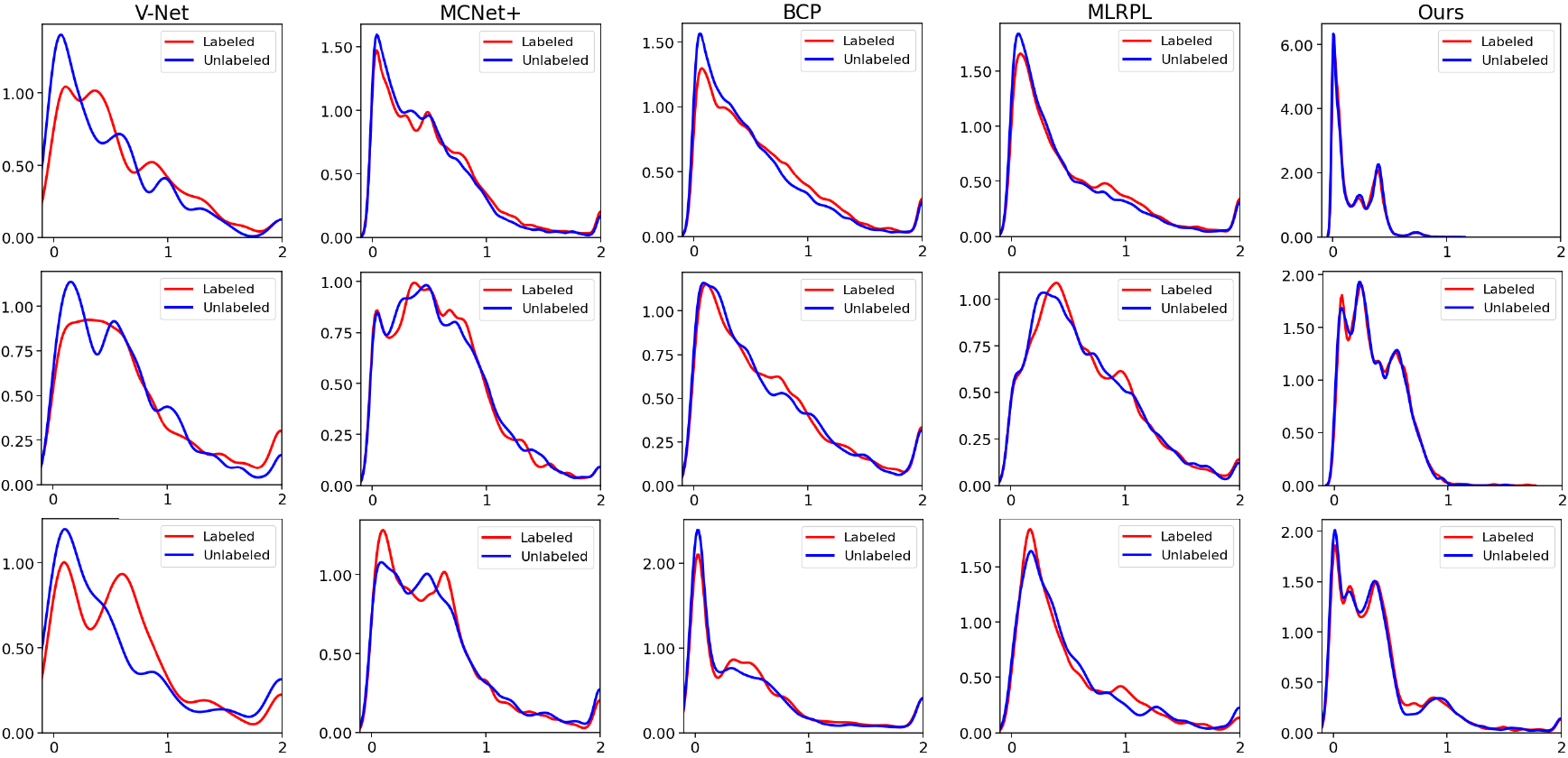}
    \caption{Kernel density estimation results from different methods, trained with 10\% labeled data on LA ($1^{st}$ row), 20\% labeled data on Pancreas ($2^{nd}$ row), and 10\% labeled data on IAN ($3^{rd}$ row). For a better comparison, the results for all methods are presented under the same feature (horizontal-axis) range of approximately 0 to 2.  
    }
    \label{fig:KDE}
\end{figure*}

\input{results/matching_score}

To assess the effect of the unlabeled-labeled distribution matching, 
we first utilize the kernel density estimation technique, introduced in BCP~\cite{bai2023bidirectional}, 
to visualize the feature distribution (e.g., histogram) of a specific class, typically the foreground. 
Differently from BCP~\cite{bai2023bidirectional} that visualizes only one single labeled and unlabeled case, 
we opt to visualize all labeled or unlabeled cases to enable a comprehensive evaluation,  
where we randomly select $1.0 \times 10^{4}$ (the same number as~\cite{bai2023bidirectional}) true positive foreground voxels from all 3D labeled or unlabeled images. 
% We compare our methods with the supervised-only V-Net~\cite{milletari2016v} and two SSS methods: MCNet+~\cite{wu2022exploring} and BCP~\cite{bai2023bidirectional} trained on LA, Pancreas, and IAN dataset with 10\%, 20\% and 10\% respectively. 
% Based on the evaluation protocol in~\cite{bai2023bidirectional}, we randomly sample $10^3$ true positive (foreground) points after evaluating all the labeled and unlabeled cases respectively, 
The results are revealed in Fig.~\ref{fig:KDE}, 
where the V-Net baseline shows a pronounced distribution gap between labeled and unlabeled samples. 
Our DuCiSC significantly reduces this gap, aligning the features of labeled and unlabeled data more effectively than other advanced semi-supervised approaches.
Interestingly, the features learned by our model fall within a relatively narrow range from 0 to 1 (particularly in LA and Pancreas datasets), compared with other approaches, 
highlighting our model's ability to extract more tightly concentrated (i.e., more similar) features across all labeled and unlabeled training images.
In order to further evaluate the matching effect quantitatively, 
we also compute the whole-set labeled-unlabeled semantic matching score $Q$ defined in Section~\ref{sec:metrics} on the LA, Pancreas, and IAN datasets.
The results in Table~\ref{tab:matching} clearly demonstrate that our proposed DuCiSC method substantially outperforms other approaches across all datasets, 
further affirming the effectiveness of our method in achieving semantic consistency between labeled and unlabeled data.

\subsection{Analytic Ablation Studies}
In this section, we perform detailed ablation experiments on LA and Pancreas datasets, to investigate the effect of each important component in our method.

\input{results/ablation}
\input{results/ablation_estimation}
\subsubsection{Intra-image Semantic Consistency}
We first study the effectiveness of the voxel-level semantic consistency strategy, with results provided in Table~\ref{tab:ablation_results}, 
where the baseline DuCiSC employs only the supervised loss $\mathcal{L}_{sup}$ on labeled data. 
We notice that incorporating $\mathcal{L}_{cs}^{ulb}$ can significantly improve the segmentation performances, 
showing the importance of enforcing the voxel-level semantic consistency between the predictions by the student and teacher model, in line with the well-established MT framework. 
A similar phenomenon can be observed with the utilization of $\mathcal{L}_{cs}^{mix}$, which also results in a notable performance boost. 
The performance gain of $\mathcal{L}_{cs}^{ulb}$ is slightly larger than that of $\mathcal{L}_{cs}^{mix}$,
because the effectiveness of mixed pseudo labels relies on the reliability of the pseudo label generated for unlabeled samples.

\subsubsection{Cross-image Semantic Consistency}
In Table~\ref{tab:ablation_results}, we also provide ablation for our proposed dual paradigms of enforcing cross-image semantic consistency via prototypes between different pairs of training images. 
To be specific, the addition of $\mathcal{L}_{proto}^{ulb}$, on the basis of $\mathcal{L}_{cs}^{ulb}$, helps the model achieve region-level semantic consistency between labeled and unlabeled training samples, 
contributing to a notable Dice improvement of 1.24\% and 1.69\% on LA and Pancreas datasets, respectively. 
Likewise, the inclusion of $\mathcal{L}_{proto}^{mix}$, in conjunction with $\mathcal{L}_{cs}^{mix}$, can bring a large improvement of approximately 1.18\% (LA) and 1.46\% (Pancreas) in terms of the Dice score metric, 
because of the semantic alignment of region-level prototypes between labeled and fused images.

\subsubsection{Self-aware Confidence Estimation}

To validate the advantage of our self-aware confidence estimation strategy in pseudo labeling, we apply various existing confidence estimation approaches to our DuCiSC method, including the probability-based~\cite{sohn2020fixmatch}, entropy-based~\cite{xu2023ambiguity}, and Monte Carlo (MC) dropout~\cite{yu2019uncertainty}.
Table~\ref{tab:ablation_estimation_results} shows the experimental results, where  we also provide the experiment without excluding unreliable voxels, denoted by Baseline, meaning that all pseudo-labeled voxels are used regardless of their confidence. 
As evident, all these confidence estimation techniques are useful in selecting reliable pseudo labels, thereby improving the segmentation performance. 
Compared with other techniques, our self-aware strategy shows more substantial improvements in all datasets, 
indicating the strength of dynamically adjusting the confidence based on the model's learning status.
To validate whether the performance gain of our self-aware strategy is statistically significant, we also conducted the statistical analysis using one-sided paired t-test for our self-aware strategy with other techniques, where the p-values associated with all pairs are below the significance level of 0.05 on the three benchmarks, statistically verifying the superiority of our proposed self-aware strategy over other techniques.

In particular, our self-aware strategy has more notable performance gain on the setting of 20\% labeled training data, compared with 10\% labeled training data. 
This observation suggests that a larger amount of labeled training data enables our model to make more accurate predictions on unlabeled training samples, contributing to a more precise estimation for our self-aware confidence thresholds. 
In addition, the advantage of our self-aware strategy is more pronounced in challenging scenarios, such as the IAN that has complex tube-like anatomical structures, outperforming other prior studies with improvements of at least 1.87\% and 3.14\% (Dice) in the two settings, respectively.

\subsubsection{Sensitivity Analysis of $\lambda_1, \lambda_2, \lambda_3$, $\lambda_4$, and $\beta$}
\begin{figure}[t!]
    \centering
    \includegraphics[width=0.5\textwidth]{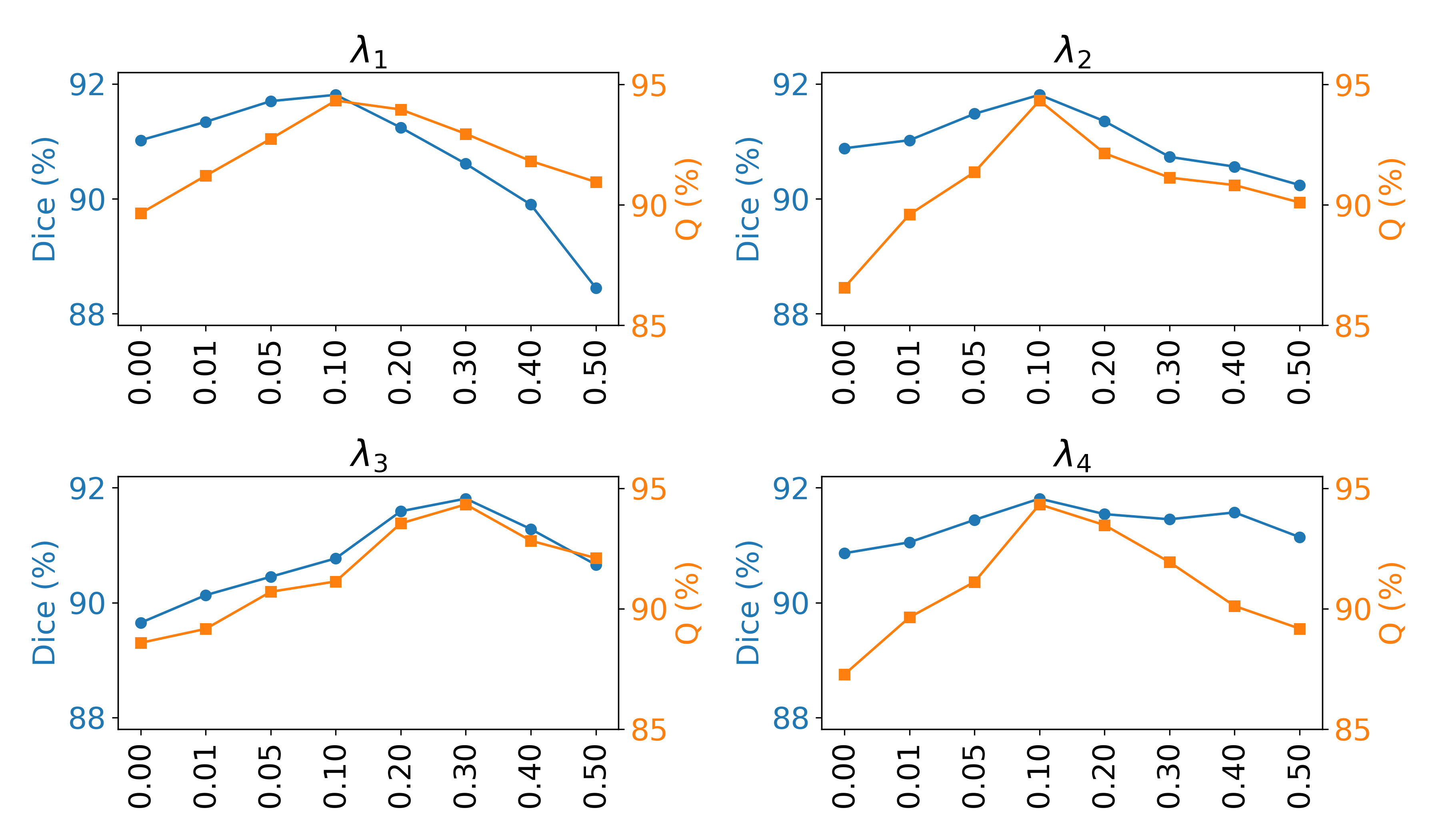}
    \caption{Sensitivity analysis of our DuCiSC method to $\lambda_1, \lambda_2, \lambda_3$, and $\lambda_4$, which is measured with both the segmentation quality (Dice score) and the alignment quality (labeled-unlabeled semantic matching score $Q$). The model is trained using 10\% labeled data on LA dataset.}
    \label{fig:lambda_ablation}
\end{figure}

We also conduct experiments to study the effect of the hyper-parameters $\lambda_1, \lambda_2, \lambda_3,$ and $\lambda_4$ in Eq.~(\ref{eq:loss}), which govern the contribution of the respective loss term, on both the segmentation quality (measured by Dice score) and the alignment quality (measured by labeled-unlabeled semantic matching score $Q$).
We vary each of them from 0.0 to 0.5 while keeping the others fixed at their current best-performing values. 
The experimental results on LA dataset are illustrated in Fig.~\ref{fig:lambda_ablation}. 
As evident, if $\lambda_1$ (controlling the labeled-unlabeled prototype alignment loss $\mathcal{L}_{proto}^{ulb}$) is too small, 
it fails to provide adequate alignment effect, leading to compromised segmentation performance. 
However, if $\lambda_1$ is too large, it can overshadow other losses (particularly the supervised loss $\mathcal{L}_{sup}$), resulting in diminished learning effectiveness. 
A similar phenomenon is also observed for $\lambda_2$, which controls the mixed-unlabeled prototype alignment loss $\mathcal{L}_{proto}^{mix}$.
Regarding $\lambda_3$ and $\lambda_4$, we find that they can consistently help achieve a better segmentation performance, 
indicating that it is important and indispensable to enforce the voxel-level semantic consistency (i.e., $\mathcal{L}_{cs}^{ulb}$ and $\mathcal{L}_{cs}^{mix}$). 
Interestingly, we observe that the segmentation quality generally follows a similar trend to the alignment quality, which further suggests that improved semantic alignment of prototypes between the labeled and unlabeled training samples contributes to better segmentation results. 
Based on Fig.~\ref{fig:lambda_ablation}, our method obtain the best performance with $\lambda_1=0.1, \lambda_2=0.1, \lambda_3=0.3,$ and $\lambda_4=0.1$. 
We also use these values in other experimental settings and datasets.

\begin{figure}[t!]
    \centering
    \includegraphics[width=0.50\textwidth]{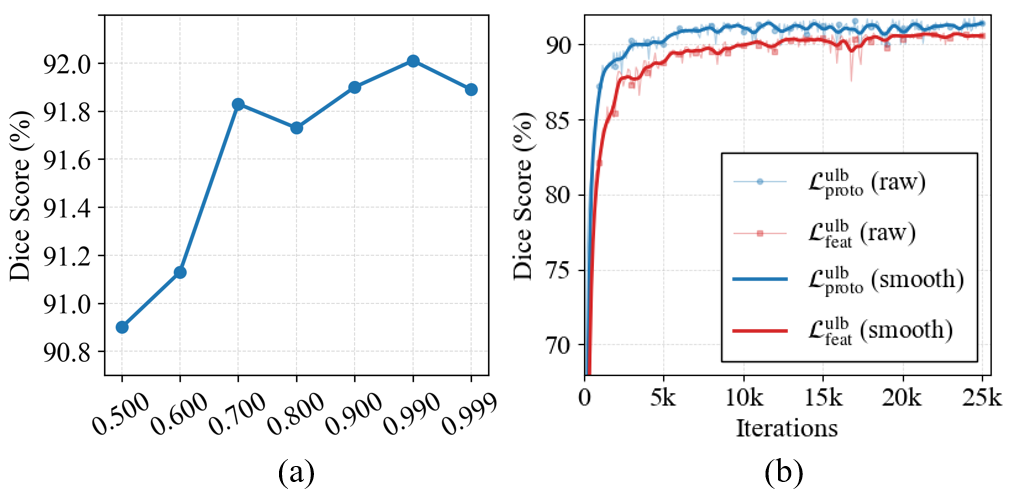}
    \caption{(a) Sensitivity analysis for the smoothing factor $\beta$ on LA dataset with 10\% labeled training data. (b) Test Dice score across training iterations on LA dataset with 10\% annotations for every 100 iterations. For better visualization, the raw curve is smoothed to illustrate the trend clearly for both approaches.}
    \label{fig:beta}
\end{figure}

Additionally, we also study the impact of $\beta$ used in the self-aware confidence estimation on segmentation performance. 
As shown in Fig.~\ref{fig:beta} (a), the segmentation results (Dice score) of our method are fairly robust when $\beta>0.9$, and are best at $\beta=0.990$. 
Therefore, we choose $\beta=0.990$ in our method.

\input{results/ablation_consistency}
\input{results/ablation_selfensembling}

\subsubsection{Choice of Mix-based Strategy}

To investigate the effect of different mix-based strategies, we have made an additional ablation study by integrating our approach with other popularly-used mixing strategies, including CutOut~\cite{devries2017improved}, CutMix~\cite{yun2019cutmix}, and Copy-Paste~\cite{ghiasi2021simple}. 
Experimental results in Table. \ref{tab:mix} illustrates that employing Mixup in our DuCiSC method achieves the best performance among the evaluated mixing strategies for our DuCiSC method, 
validating our choice.

\input{results/MixCompare}

\section{Discussion and future works}

The core of our DuCiSC approach lies in the use of class-specific prototypes, which serve as the foundation for establishing prototype alignments between paired training samples: 1) labeled and unlabeled images and 2) labeled and fused images, as in Eq.~(\ref{proto_unlabled}) and Eq.~(\ref{eq:prototype_fused}). 
A natural question may arise about how our approach compares to direct feature-based consistency methods~\cite{you2022simcvd,zhang2023semi,su2024consistency,huang2024dual}, which minimize the discrepancy between features extracted by the student and teacher networks. 
When adapted to our framework, these methods would reformulate Eq.~(\ref{proto_unlabled}) and Eq.~(\ref{eq:prototype_fused}) as: 
\begin{equation}
\begin{aligned}
    \mathcal{L}^{ulb}_{feat} = \frac{1}{H\times^s W^s \times D^s} \sum_{s=1}^S || \mathbf{F}_j^{s, tea} - \mathbf{F}_j^{s, stu}||^2_F, \\
    \mathcal{L}^{mix}_{feat} = \frac{1}{H\times^s W^s \times D^s} \sum_{s=1}^S || \mathbf{F}_k^{s, tea} - \mathbf{F}_k^{s, stu}||^2_F,
\end{aligned}
\label{eq:feat_align}
\end{equation}
where $||\cdot||^2_F$ represents Frobenius norm, $\mathbf{F}_j^{s, tea}$ and $\mathbf{F}_j^{s, stu}$ denote the features (at scale $s$) extracted from an unlabeled image $\mathbf{x}_j$ by the teacher and student models, respectively. 
Notice that, these features are taken from the same layer used for computing our prototypes to ensure comparison fairness. 
The comparative results in Table~\ref{tab:ablation_feature_based} demonstrate that our approach, which aligns class-specific prototypes, 
substantially outperforms the class-agnostic direct feature-based consistency. 
To understand this advantage more clearly, we provide a theoretical analysis.
We first derive the gradient of the direct feature consistency loss $\mathcal{L}^{ulb}_{feat}$ (omitting scale $s$ for simplicity):
\begin{equation}
    \label{eq:feature_grad}
    \nabla_{\theta_{}}\mathcal{L}_{feat}^{ulb} = \frac{2}{H\times W \times D} \sum_{h, w, d}^{H, W, D} \mathbf{F}_{j}^{tea}(h,w,d) - \mathbf{F}_{j}^{stu}(h,w,d).
\end{equation}
In contrast, according to Eq.~(\ref{proto_unlabled}), the gradient of our cross-image prototype alignment loss $\mathcal{L}_{proto}^{ulb}$ is:
\begin{equation}
\begin{split}
    \nabla_{\theta_{}}\mathcal{L}_{proto}^{ulb} &  =  \frac{2}{C}\sum_{c=1}^C \mathbf{p}_i^c-\mathbf{p}_j^c \\
    & = \frac{2}{C}\sum_{c=1}^C \frac{ \sum_{(h, w, d)}^{(H, W, D)} \mathbf{y}^c_i(h, w, d) \cdot \mathbf{F}_i(h, w, d)}{\sum_{(h, w, d)}^{(H, W, D)} \mathbf{y}^c_i(h, w, d)} -\\
    & \qquad \ \quad \quad \frac{ \sum_{(h, w, d)}^{(H, W, D)} \bar{\mathbf{y}}^c_j(h, w, d) \cdot \mathbf{F}_j(h, w, d)}{\sum_{(h, w, d)}^{(H, W, D)} \bar{\mathbf{y}}^c_j(h, w, d)}. 
\end{split}
\label{eq:grad_proto_unlabled}
\end{equation}
While $\mathcal{L}_{feat}^{ulb}$ and $\mathcal{L}_{proto}^{ulb}$, as well as their gradients, exhibit similar mathematical structures, 
they differ fundamentally in two important ways:
% that make our cross-image prototype consistency approach yields stronger penalty and fine-grained feature alignment during training.
1) Cross-image prototype alignment introduces stronger penalties.
As shown in Eq. (\ref{eq:feat_align}) and Eq. (\ref{eq:feature_grad}), the direct feature-based loss $\mathcal{L}_{feat}^{ulb}$ relies on comparing teacher and student features from the same input $\mathbf{x}_j$. 
While effective in early training, this feature discrepancy diminishes as the models become more stable, weakening the supervisory signal.
In contrast, our prototype alignment loss $\mathcal{L}_{proto}^{ulb}$ compares features from different inputs, i.e., $\mathbf{x}_i$ and $\mathbf{x}_j$, maintaining a non-trivial discrepancy throughout training. This cross-image setting continually provides a stronger and more diverse supervisory signal, especially beneficial in the later stages of training;
2) Class-wise prototype alignment enables fine-grained supervision.
Unlike direct feature-based methods that perform class-agnostic consistency, 
our method performs alignment of prototypes (averaged features) in a class-wise manner, as in Eq. (\ref{proto_unlabled}) and Eq. (\ref{eq:grad_proto_unlabled}).
This design enables fine-grained feature consistency between samples.
For example, if there exists a significant feature discrepancy between labeled and unlabeled training samples for a particular class, 
that class will dominate both the loss $\mathcal{L}_{proto}^{ulb}$ and the gradient $\nabla_{\theta_{}}\mathcal{L}_{proto}^{ulb}$. 
As a result, the model receives a uniquely high penalty signal for that specific class, driving targeted and class-sensitive feature consistency.
To corroborate this analysis, we also visualize the learning dynamics in Fig.~\ref{fig:beta} (b), which compares the test Dice scores over training iterations for both approaches. 
Our prototype-based method demonstrates faster convergence and higher final performance, likely due to the stronger and more informative training signals it provides.

\begin{figure}[ht]
    \centering
    \includegraphics[width=0.485\textwidth]{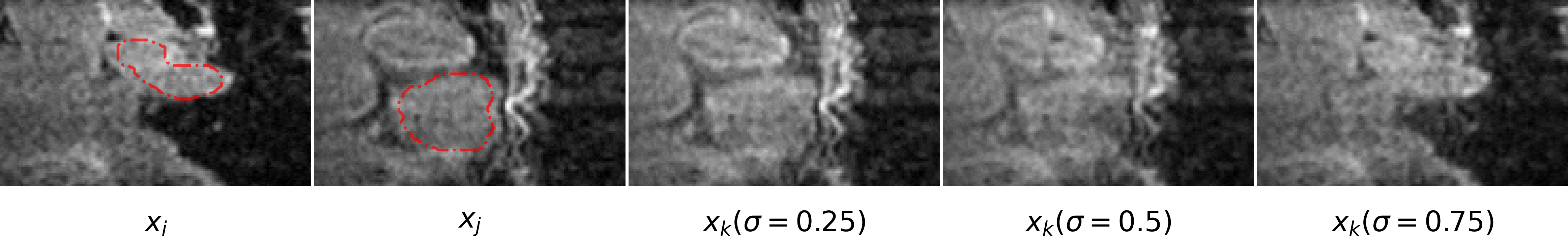}
    \caption{Illustration of the mixed images with different combination ratios used in our method, where the red dashed contour denotes the ground-truth LA boundaries and $\sigma$ is the combination ratio. We observe that the LA boundaries in $\mathbf{x}_j$ are barely visible in the mixed image $\mathbf{x}_k$ (e.g., $\sigma$ = 0.75), which makes the segmentation on them particularly challenging.}
    \label{fig:compare_bcp}
\end{figure}

Our approach is built upon the Mean Teacher, 
a popular self-ensembling framework that creates a robust teacher model by averaging the student model's weights using an EMA strategy. 
One may have concerns on adapting our method to other self-ensembling mechanisms, e.g., Temporal Ensembling~\cite{laine2016temporal} that maintains an EMA prediction at the data level for each training sample.
To achieve this, we adapt our method by employing the student network itself to generate pseudo-label predictions, which are updated in an EMA manner per 10 epochs. Other key innovations in our approach (such as dual paradigms of cross-image prototype alignment and self-aware confidence estimation) remain unchanged.
The experimental results are presented in Table~\ref{tab:self-ensembling}, 
where the Mean Teacher strategy achieves better segmentation results than the Temporal Ensembling, in line with findings observed in the established semi-supervised learning methods~\cite{tarvainen2017mean,cui2019semi}. 
We attribute this to the fact that while Temporal Ensembling improves the quality of pseudo-label predictions for unlabeled training samples, its infrequent updates (once every several epochs) limit training efficiency and overall effectiveness.

Although our method effectively enforces region-level semantic consistency between labeled and unlabeled training images using prototypes, achieving precise alignment for more localized structures remains challenging, particularly for small and highly deformable anatomical regions, such as the pancreatic tail or alveolar nerve. 
We believe that further investigation is needed to ensure fine-grained semantic alignment across different regions within the same organ between labeled and unlabeled training images. 
In addition, our method relies on the mixup strategy to fuse the labeled and unlabeled training images, as illustrated in Fig.~\ref{fig:compare_bcp}, where the resulting mixed images can exhibit significant changes in anatomical features depending on the combination ratio.

\section{Conclusion}
\label{sec:conclusion}
In this paper, we proposed the effective DuCiSC framework for semi-supervised medical image segmentation. The DuCiSC method leverages dual paradigms to enforce the consistency of region-level semantics by aligning the prototypes of labeled images with unlabeled and fused images, which effectively overcome the distribution mismatch issue. Moreover, a self-aware confidence estimation approach is introduced to flexibly identify highly-reliable voxels in unlabeled training images, allowing the model to leverage the unlabeled data according to its learning status.
The extensive experimental results on four public benchmarks verified the superiority and robustness of our method over existing state-of-the-art approaches.

{
\bibliographystyle{ieeetr}
\bibliography{reference}
}

\end{document}

%% file: results/LA_new.tex
\begin{table}[!h]
    \centering\caption{Segmentation performance comparison with other competing approaches on LA dataset,
    with best results highlighted in bold and second best results marked in underline. 
    }
    \label{tab:la_results}
    \setlength{\tabcolsep}{1.4 mm}
    \resizebox{0.999\linewidth}{!}{
    \begin{tabular}{c|cc|cccc}
    \hline
    \multirow{2}{*}{\begin{tabular}[c]{@{}c@{}} \\ Method\end{tabular}}                         & \multicolumn{2}{c|}{Data used} & \multicolumn{4}{c}{Metrics}  \\ \cline{2-7} 
        &
        Labeled &
        Unlabeled &
        \begin{tabular}[c]{@{}c@{}}Dice\\ (\%)↑\end{tabular} &
        \begin{tabular}[c]{@{}c@{}}Jaccard\\ (\%)↑\end{tabular} &
        \begin{tabular}[c]{@{}c@{}}95HD\\ (voxel)↓\end{tabular} &
        \begin{tabular}[c]{@{}c@{}}ASD\\ (voxel)↓\end{tabular} \\ \hline
    V-Net & 8 (10\%)                   & 0              & 79.53 & 67.66 & 24.23 & 7.83 \\
    V-Net & 16 (20\%)                  & 0              & 84.93 & 75.87 & 14.50 & 4.30 \\
    V-Net & 80 (100\%)                 & 0              & 91.33 & 84.62 & 8.56  & 2.20 \\ \hline
    UA-MT~\cite{yu2019uncertainty}     &                &                & 84.58 & 73.77 & 18.76 & 4.90 \\
    DTC~\cite{luo2021semi}             &                &                & 85.32 & 74.93 & 11.42 & 2.37 \\
    CPCL~\cite{xu2022all}              &                &                & 86.20 & 76.00 & 11.43 & 2.52 \\
    MC-Net~\cite{wu2021semi}           &                &                & 87.27 & 78.17 & 11.14 & 2.34 \\
    SCP-Net~\cite{zhang2023self}       &                &                & 87.68 & 78.89 & 10.98 & 2.28 \\
    \textcolor{black}{FixMatch~\cite{sohn2020fixmatch}}   &                &                & \textcolor{black}{87.79} & \textcolor{black}{78.33} & \textcolor{black}{9.42}  & \textcolor{black}{2.44} \\
    MC-Net+~\cite{wu2022mutual}        &                &                & 88.39 & 79.22 & 8.34  & 1.87 \\
    SimCVD~\cite{you2022simcvd}        &                &                & 89.03 & 80.34 & 8.34  & 2.59 \\
    CAML~\cite{gao2023correlation}     &  8 (10\%)      &  72 (90\%)     & 89.62 & 81.28 & 8.76  & 2.02 \\
    \textcolor{black}{UniMatch~\cite{yang2023revisiting}} &                &                & \textcolor{black}{89.09} & \textcolor{black}{80.47} & \textcolor{black}{12.50} & \textcolor{black}{3.59} \\
    \textcolor{black}{PS-MT~\cite{liu2022perturbed}}      &                &                & \textcolor{black}{89.72} & \textcolor{black}{81.48} & \textcolor{black}{6.94}  & \textcolor{black}{1.92} \\
    BCP~\cite{bai2023bidirectional}    &                &                & 89.09 & 80.49 & 7.49  & 1.95 \\
    EIC~\cite{huang2024exploring}      &                &                & 89.25 & 80.68 & 6.96  & 1.86 \\
    \textcolor{black}{TraCoCo~\cite{liu2024translation}} &                &                & \textcolor{black}{89.29} & \textcolor{black}{80.82} & \textcolor{black}{6.92}  & \textcolor{black}{2.28} \\
    MLRPL~\cite{su2024mutual}          &                &                & 89.86 & 81.68 & 6.91  & 1.85 \\
    \textcolor{black}{AD-MT~\cite{zhao2024alternate}}     &                &                & \textcolor{black}{90.55} & \textcolor{black}{82.79} & \textcolor{black}{5.81}  & \textcolor{black}{\underline{1.70}} \\
    \textcolor{black}{DistillMatch~\cite{wang2024distillmatch}}&           &                & \textcolor{black}{\underline{90.58}} & \textcolor{black}{\underline{82.86}} & \textcolor{black}{\underline{5.42}}  & \textcolor{black}{1.82} \\
    DuCiSC                             &                &                & \textbf{91.81} & \textbf{84.93} & \textbf{5.08}  & \textbf{1.54} \\  \hline
    UA-MT~\cite{yu2019uncertainty}     &                &                & 87.30 & 78.06 & 9.72  & 2.60 \\
    DTC~\cite{luo2021semi}             &                &                & 88.32 & 79.34 & 8.72  & 2.02 \\
    CPCL~\cite{xu2022all}              &                &                & 87.68 & 79.20 & 9.13  & 2.13 \\
    \textcolor{black}{FixMatch~\cite{sohn2020fixmatch}}   &                &                & \textcolor{black}{90.33} & \textcolor{black}{82.43} & \textcolor{black}{6.36}  & \textcolor{black}{1.64} \\
    MC-Net~\cite{wu2021semi}           &                &                & 90.43 & 82.81 & 6.58  & 1.60 \\
    SCP-Net~\cite{zhang2023self}       &                &                & 90.41 & 81.87 & 6.59  & 1.86 \\
    MC-Net+~\cite{wu2022mutual}        &                &                & 90.58 & 82.87 & 6.35  & 1.56 \\
    SimCVD~\cite{you2022simcvd}        &  16 (20\%)     &  64 (80\%)     & 90.85 & 83.80 & 6.03  & 1.86 \\
    CAML~\cite{gao2023correlation}     &                &                & 90.78 & 83.19 & 6.11  & 1.68 \\  
    \textcolor{black}{UniMatch~\cite{yang2023revisiting}} &                &                & \textcolor{black}{90.77} & \textcolor{black}{83.18} & \textcolor{black}{7.21}  & \textcolor{black}{2.05} \\
    BCP~\cite{bai2023bidirectional}    &                &                & 90.38 & 82.57 & 6.68  & 1.76 \\
    \textcolor{black}{TraCoCo~\cite{liu2024translation}} &                &                & \textcolor{black}{90.94} & \textcolor{black}{83.47} & \textcolor{black}{5.49}  & \textcolor{black}{1.79} \\
    MLRPL~\cite{su2024mutual}          &                &                & 91.02 & 83.62 & 5.78  & 1.66 \\
    \textcolor{black}{DistillMatch~\cite{wang2024distillmatch}}&           &                & \textcolor{black}{\underline{91.59}} & \textcolor{black}{\underline{84.54}} & \textcolor{black}{\underline{5.23}}  & \textcolor{black}{\underline{1.48}} \\
    DuCiSC                             &                &                & \textbf{92.11} & \textbf{85.42} & \textbf{4.98}  & \textbf{1.36} \\ \hline
    \end{tabular}%
    }
    \end{table}

%% file: results/Pancreas_new.tex
\begin{table}[h]
    \centering\caption{Performance comparison with other competing approaches on NIH-Pancreas,
    with best results highlighted in bold and second best results marked in underline.
    % "↑" and "↓" indicate the larger and the smaller the better.
    }
    \label{tab:pan_results}
    \setlength{\tabcolsep}{1.5 mm}
    \resizebox{0.999\linewidth}{!}{%
    \begin{tabular}{c|cc|cccc}
    \hline
    \multirow{2}{*}{\begin{tabular}[c]{@{}c@{}} \\ Method\end{tabular}}       & \multicolumn{2}{c|}{Data used}  & \multicolumn{4}{c}{Metrics}          \\ \cline{2-7} 
                & Labeled                   & Unlabeled                 & \begin{tabular}[c]{@{}c@{}}Dice\\ (\%)↑\end{tabular} & \begin{tabular}[c]{@{}c@{}}Jaccard\\ (\%)↑\end{tabular} & \begin{tabular}[c]{@{}c@{}}95HD\\ (voxel)↓\end{tabular} & \begin{tabular}[c]{@{}c@{}}ASD\\ (voxel)↓\end{tabular} \\ \hline
    V-Net       & 6 (10\%)                   & 0    & 54.94      & 40.87         & 47.48         & 17.43        \\
    V-Net       & 12 (20\%)                  & 0    & 71.52      & 57.68         & 18.12         & 5.41         \\
    V-Net       & 62 (100\%)                   & 0    & 82.60      & 70.81         & 5.61          & 1.33         \\ \hline
    UA-MT~\cite{yu2019uncertainty}  & \multirow{10}{*}{6 (10\%)}  & \multirow{10}{*}{56 (90\%)} & 66.44      & 52.02         & 17.04         & 3.03         \\
    DTC~\cite{luo2021semi}          &      &      & 66.58      & 51.79         & 15.46         & 4.16         \\
    MC-Net~\cite{wu2021semi}        &      &      & 69.07      & 54.36         & 14.53         & 2.28         \\
    MC-Net+~\cite{wu2022mutual}     &      &      & 70.00      & 55.66         & 16.03         & 3.87         \\
    \textcolor{black}{UniMatch~\cite{yang2023revisiting}} &    &      & \textcolor{black}{69.90}      & \textcolor{black}{55.13}         & \textcolor{black}{12.94}         & \textcolor{black}{3.56}    \\
    \textcolor{black}{PS-MT~\cite{liu2022perturbed}}   &      &      & \textcolor{black}{76.94}      & \textcolor{black}{62.37}         & \textcolor{black}{13.12}         & \textcolor{black}{3.66}         \\
    MLRPL~\cite{su2024mutual}       &      &      & 75.93      & 62.12         & 9.07          & \underline{1.54}         \\
    \textcolor{black}{TraCoCo~\cite{liu2024translation}} &                &                & \textcolor{black}{79.22} & \textcolor{black}{66.04} & \textcolor{black}{8.46}  & \textcolor{black}{2.57} \\
    \textcolor{black}{AD-MT~\cite{zhao2024alternate}}  &      &      & \textcolor{black}{\underline{80.21}}      & \textcolor{black}{\underline{67.51}}         & \textcolor{black}{\textbf{7.18}}          & \textcolor{black}{1.66}    \\
    DuCiSC      &      &      & \textbf{80.72}            & \textbf{68.12}                 & \underline{7.20}                 & \textbf{1.53}                \\ \hline
    UA-MT~\cite{yu2019uncertainty}  & \multirow{12}{*}{12 (20\%)} & \multirow{12}{*}{50 (80\%)} & 76.10      & 62.62         & 10.84         & 2.43         \\
    DTC~\cite{luo2021semi}          &      &      & 76.27      & 62.82         & 8.70          & 2.20         \\
    MC-Net~\cite{wu2021semi}        &      &      & 78.17      & 65.22         & 6.90          & 1.55         \\
    MC-Net+~\cite{wu2022mutual}     &      &      & 79.37      & 66.83         & 8.52          & 1.72         \\
    SimCVD~\cite{you2022simcvd}     &      &      & 75.39      & 61.56         & 9.84          & 2.33         \\
    \textcolor{black}{UniMatch~\cite{yang2023revisiting}} &    &      & \textcolor{black}{79.52}      & \textcolor{black}{66.64}         & \textcolor{black}{13.05}         & \textcolor{black}{3.02}    \\
    \textcolor{black}{PS-MT~\cite{liu2022perturbed}}   &      &      & \textcolor{black}{80.74}      & \textcolor{black}{68.15}         & \textcolor{black}{7.41}          & \textcolor{black}{2.06}         \\
    EIC~\cite{huang2024exploring}   &      &      & 81.17      & 68.68         & 6.17          & 1.46         \\
    MLRPL~\cite{su2024mutual}       &      &      & 81.53      & 69.35         & 6.81          & \underline{1.33}         \\
    \textcolor{black}{TraCoCo~\cite{liu2024translation}} &                &                & \textcolor{black}{81.80} & \textcolor{black}{69.56} & \textcolor{black}{5.70}  & \textcolor{black}{1.49} \\
    \textcolor{black}{AD-MT~\cite{zhao2024alternate}}  &      &      & \textcolor{black}{82.61}      & \textcolor{black}{70.70}         & \textcolor{black}{\underline{4.94}}          & \textcolor{black}{1.38}    \\
    BCP~\cite{bai2023bidirectional} &      &      & \underline{82.91}      & \underline{70.97}         & 6.43          & 2.25         \\
    DuCiSC       &      &      & \textbf{83.71}            & \textbf{72.29}                 & \textbf{4.46}                  & \textbf{1.32}                 \\ \hline
    \end{tabular}%
    }
    \end{table}

%% file: results/Nerves_results.tex
% \usepackage{graphicx}
\begin{table}[h]
\centering\caption{Segmentation performance comparison with other competing approaches on IAN dataset,
with best results highlighted in bold and second best results marked in underline.
% "↑" and "↓" indicate the larger and the smaller the better.
}
\label{tab:nerves_results}
\setlength{\tabcolsep}{1.2 mm}
\resizebox{1.0\linewidth}{!}{%
\begin{tabular}{c|cc|cccc}
\hline
\multirow{2}{*}{\begin{tabular}[c]{@{}c@{}} \\ Method\end{tabular}}           & \multicolumn{2}{c|}{Data used} & \multicolumn{4}{c}{Metrics}  \\ \cline{2-7} 
 &
  Labeled &
  Unlabeled &
  \begin{tabular}[c]{@{}c@{}}Dice\\ (\%)↑\end{tabular} &
  \begin{tabular}[c]{@{}c@{}}Jaccard\\ (\%)↑\end{tabular} &
  \begin{tabular}[c]{@{}c@{}}95HD\\ (voxel)↓\end{tabular} &
  \begin{tabular}[c]{@{}c@{}}ASD\\ (voxel)↓\end{tabular} \\ \hline
V-Net           & 9 (10\%)        & 0 & 58.52 & 43.81 & 28.63 & 6.65 \\
V-Net           & 18 (20\%)       & 0 & 71.09 & 56.66 & 15.91 & 4.02 \\
V-Net           & 90 (100\%)      & 0 & 78.80 & 65.73 & 14.86 & 3.52 \\ \hline
UA-MT~\cite{yu2019uncertainty}    &           &             & 63.39 & 48.78 & 21.06 & 4.45 \\ 
BCP~\cite{bai2023bidirectional}   &           &             & 68.27 & 54.05 & 18.91 & 4.82 \\
MC-Net+~\cite{wu2022mutual}       &  9 (10\%) & 81 (90\%)   & 70.95 & 56.59 & \underline{18.73} & \underline{4.41} \\
MLRPL~\cite{su2024mutual}         &           &             & \underline{71.10} & \underline{56.74} & 21.33 & 4.96 \\
DuCiSC      &   &   & \textbf{76.79} & \textbf{63.00} & \textbf{16.39} & \textbf{3.84} \\ \hline
UA-MT~\cite{yu2019uncertainty}  &        &        & 72.59 & 57.98 & 15.74 & 3.90 \\ 
BCP~\cite{bai2023bidirectional} &        &        & 74.82 & 60.62 & 15.37 & 4.21 \\
MC-Net+~\cite{wu2022mutual}     & 18 (20\%)       & 72 (80\%)   & 76.18 & \underline{62.65} & 14.06 & \underline{3.26} \\
MLRPL~\cite{su2024mutual}       &   &   & \underline{76.28} & 62.43 & \underline{13.41} & 3.31 \\
% DuCiSC       &   &   & \textbf{78.69} & \textbf{65.75} & \underline{13.93} & \textbf{3.16} \\ \hline
DuCiSC       &   &   & \textbf{78.41} & \textbf{65.22} & \textbf{13.37} & \textbf{3.23} \\ \hline
\end{tabular}%
}
\end{table}

%% file: results/ACDC.tex
\begin{table}[!h]
    \centering\caption{Segmentation performance among different approaches on ACDC with 10\% labeled training data.} 
    \label{tab:ACDC}
    \setlength{\tabcolsep}{1.5 mm}
    \resizebox{\linewidth}{!}{%
    \begin{tabular}{c|cc|cccc}
        \hline
        \multirow{2}{*}{Method}        & \multicolumn{2}{c|}{Date Used} & \multicolumn{4}{c}{Metrics} \\ \cline{2-7} 
     & Labeled        & Unlabeled     & \begin{tabular}[c]{@{}c@{}}Dice\\ (\%)↑\end{tabular} & \begin{tabular}[c]{@{}c@{}}Jaccard\\ (\%)↑\end{tabular} & \begin{tabular}[c]{@{}c@{}}95HD\\ (voxel)↓\end{tabular} & \begin{tabular}[c]{@{}c@{}}ASD\\ (voxel)↓\end{tabular} \\ \hline
        U-Net     & 7 (10\%)       & 0               & 79.41    & 68.11       & 9.35        & 2.70       \\
        U-Net     & 70 (100\%)      & 0              & 91.44    & 84.59       & 4.30        & 0.99       \\ \hline
        UA-MT~\cite{yu2019uncertainty} &      &     & 81.65    & 70.64       & 6.88        & 2.02       \\
        URPC~\cite{luo2021efficient}   &      &     & 83.10    & 72.41       & 4.84        & 1.53       \\
        SASSNet~\cite{li2020shape}     &      &     & 84.50    & 73.34       & 5.42        & 1.86       \\
        DTC~\cite{luo2021semi}         &      &     & 84.29    & 73.92       & 12.81       & 4.01       \\
        CPS~\cite{chen2021semi}        &      &     & 86.78    & 77.67       & 6.07        & 1.40       \\
        SS-Net~\cite{wu2022exploring}  &        &     & 86.78  & 77.67       & 6.07        & 1.40       \\
        MC-Net+~\cite{wu2022mutual}    & 7 (10\%) & 63 (90\%)     & 87.10    & 78.06       & 6.68        & 2.00       \\
        UniMatch~\cite{yang2023revisiting}       &      &     & 88.08    & 80.10       & 2.09        & 0.45       \\
        PS-MT~\cite{liu2022perturbed}  &      &     & 88.91    & 80.79       & 4.96        & 1.83       \\
        BCP~\cite{bai2023bidirectional}&      &     & 88.84    & 80.62       & 3.98        & 1.17       \\
        AD-MT~\cite{zhao2024alternate} &      &     & 89.46    & 81.47       & 1.51        & 0.44       \\
        DistillMatch~\cite{wang2024distillmatch} &      &     & \underline{89.48}      & \underline{82.00}         & \underline{1.48}& \underline{0.38}         \\
        DuCiSC   &      &     & \textbf{89.82}       & \textbf{82.28}& \textbf{1.33} & \textbf{0.38}\\ \hline
        \end{tabular}
    }
\end{table}

%% file: results/matching_score.tex
\begin{table}[h]
\centering\caption{Semantic matching score $Q$ of our DuCiSC and other leading methods: V-Net~\cite{milletari2016v}, BCP~\cite{bai2023bidirectional}, MCNet+~\cite{wu2022mutual}, and MLRPL~\cite{huang2024exploring}. }
\label{tab:matching}
\setlength{\tabcolsep}{2. mm}
\resizebox{0.85\linewidth}{!}{%
\begin{tabular}{c|c|c|c}
\hline
Method                                 & LA (10\%)          & Pancreas (20\%)    & IAN (10\%)          \\ \hline
V-Net\cite{milletari2016v}             & 0.3427             & 0.5917             & 0.3988              \\
BCP\cite{bai2023bidirectional}         & 0.4338             & 0.6300             & \underline{0.4800}              \\
MCNet+\cite{wu2022mutual}              & \underline{0.4689} & 0.7438             & 0.4469  \\
MLRPL\cite{huang2024exploring}         & 0.4292             & \underline{0.7452} & 0.4661                    \\
DuCiSC                                 & \textbf{0.9433}    & \textbf{0.9813}    & \textbf{0.8683}     \\ \hline
\end{tabular}%
}
\end{table}

%% file: results/ablation.tex
\begin{table*}[t]
\centering\caption{Ablation analysis of our DuCiSC method, experimented on LA and Pancreas datasets with 10\% labeled data.}
\label{tab:ablation_results}
\setlength{\tabcolsep}{2.4 mm}
\resizebox{\linewidth}{!}
{%
\begin{tabular}{c|cc|cc|cccc|cccc}
\hline
Supervised & \multicolumn{2}{c|}{Intra-image} & \multicolumn{2}{c|}{Cross-image} & \multicolumn{4}{c|}{LA (10\%)} & \multicolumn{4}{c}{Pancreas (10\%)} \\ \hline
$\mathcal{L}_{sup}$ &
  $\mathcal{L}_{cs}^{ulb}$ &
  $\mathcal{L}_{cs}^{mix}$ &
  $\mathcal{L}_{proto}^{ulb}$ &
  $\mathcal{L}_{proto}^{mix}$ &
  \begin{tabular}[c]{@{}c@{}}Dice\\ (\%)↑\end{tabular} &
  \begin{tabular}[c]{@{}c@{}}Jaccard\\ (\%)↑\end{tabular} &
  \begin{tabular}[c]{@{}c@{}}95HD\\ (voxel)↓\end{tabular} &
  \begin{tabular}[c]{@{}c@{}}ASD\\ (voxel)↓\end{tabular} &
  \begin{tabular}[c]{@{}c@{}}Dice\\ (\%)↑\end{tabular} &
  \begin{tabular}[c]{@{}c@{}}Jaccard\\ (\%)↑\end{tabular} &
  \begin{tabular}[c]{@{}c@{}}95HD\\ (voxel)↓\end{tabular} &
  \begin{tabular}[c]{@{}c@{}}ASD\\ (voxel)↓\end{tabular} \\ \hline
$\checkmark$ &              &              &               &              & 86.27   & 77.81   & 8.69   & 2.58   & 72.94 & 59.56 & 18.05 & 3.84    \\ 
$\checkmark$ & $\checkmark$ &              &               &              & 90.02   & 82.14   & 7.04   & 2.17   & 78.46 & 65.31 & 10.32 & 3.08    \\ 
$\checkmark$ & $\checkmark$ &              & $\checkmark$  &              & 91.26   & 83.99   & 5.63   & 1.73   & 80.15 & 68.03 & 7.28  & 1.99    \\  
$\checkmark$ &              & $\checkmark$ &               &              & 88.82   & 80.64   & 6.65   & 2.24   & 77.56 & 64.28 & 8.55  & 2.10     \\
$\checkmark$ &              & $\checkmark$ &               & $\checkmark$ & 90.00   & 81.91   & 6.01   & 1.96   & 79.02 & 65.81 & 7.65  & 2.04     \\
$\checkmark$ & $\checkmark$ & $\checkmark$ & $\checkmark$  & $\checkmark$ & 91.81   & 84.93   & 5.08   & 1.54   & 80.72 & 68.12 & 7.20  & 1.53   \\ \hline %
\end{tabular}%
}
\end{table*}

%% file: results/ablation_estimation.tex
% \usepackage{graphicx}
\begin{table*}[ht]
\centering\caption{Ablation study for our self-aware confidence estimation, conducted on LA, Pancreas, and IAN with both 10\% and 20\% labeled training data.}
    \label{tab:ablation_estimation_results}
    \setlength{\tabcolsep}{0.5 mm}
     \resizebox{1.0\linewidth}{!}{%
    \color{black}\begin{tabular}{c|cccc|cccc|cccc}
\hline
\multirow{3}{*}{Method} & \multicolumn{4}{c|}{LA}                                               & \multicolumn{4}{c|}{Pancreas}                                         & \multicolumn{4}{c}{IAN}                                              \\ \cline{2-13} 
                        & \multicolumn{2}{c|}{10\%}                 & \multicolumn{2}{c|}{20\%} & \multicolumn{2}{c|}{10\%}                 & \multicolumn{2}{c|}{20\%} & \multicolumn{2}{c|}{10\%}                 & \multicolumn{2}{c}{20\%} \\ \cline{2-13} 
                        & Dice (\%)↑ & \multicolumn{1}{c|}{p-value} & Dice (\%)↑    & p-value   & Dice (\%)↑ & \multicolumn{1}{c|}{p-value} & Dice (\%)↑    & p-value   & Dice (\%)↑ & \multicolumn{1}{c|}{p-value} & Dice (\%)↑   & p-value   \\ \hline
Baseline                & 90.55      & \multicolumn{1}{c|}{$1.06\times10^{-5}$} & 90.67         & $7.14\times10^{-5}$   & 79.69      & \multicolumn{1}{c|}{$1.11\times10^{-3}$} & 81.64         & $4.28\times10^{-2}$   & 73.01      & \multicolumn{1}{c|}{$2.58\times10^{-4}$} & 74.03        & $1.67\times10^{-6}$   \\
Probability-based~\cite{sohn2020fixmatch}      & 91.24      & \multicolumn{1}{c|}{$1.69\times10^{-3}$} & 91.36         & $1.43\times10^{-3}$   & 80.30      & \multicolumn{1}{c|}{$3.94\times10^{-3}$} & 81.90         & $3.61\times10^{-3}$   & 74.44      & \multicolumn{1}{c|}{$2.89\times10^{-5}$} & 74.58        & $4.39\times10^{-4}$   \\
Entropy-based~\cite{xu2023ambiguity}           & 91.45      & \multicolumn{1}{c|}{$4.56\times10^{-2}$} & 91.45         & $1.81\times10^{-2}$   & 80.33      & \multicolumn{1}{c|}{$4.11\times10^{-3}$} & 81.92         & $4.39\times10^{-2}$   & 74.91      & \multicolumn{1}{c|}{$9.03\times10^{-3}$} & 75.05        & $2.74\times10^{-4}$   \\
MC-Dropout~\cite{yu2019uncertainty}              & 91.42      & \multicolumn{1}{c|}{$2.59\times10^{-2}$} & 91.47         & $1.39\times10^{-2}$   & 80.52      & \multicolumn{1}{c|}{$4.91\times10^{-3}$} & 82.19         & $4.61\times10^{-2}$   & 74.92      & \multicolumn{1}{c|}{$5.98\times10^{-3}$} & 75.27        & $4.12\times10^{-6}$   \\
Ours                    & 91.81      & \multicolumn{1}{c|}{-}       & 92.11         & -         & 80.72      & \multicolumn{1}{c|}{-}       & 83.71         & -         & 76.79      & \multicolumn{1}{c|}{-}       & 78.41        & -         \\ \hline
\end{tabular}}
\end{table*}

%% file: results/ablation_consistency.tex
\begin{table*}[!h]
    \centering\caption{Segmentation performance comparison between our class-specific alignment of prototypes and the class-agnostic feature-based consistency, experimented on LA dataset.}
    \label{tab:ablation_feature_based}
    \setlength{\tabcolsep}{1.6 mm}
    \resizebox{1.0\linewidth}{!}{%
\color{black}\begin{tabular}{c|cccc|cccc}
\hline
\multirow{2}{*}{Consistency} & \multicolumn{4}{c|}{10\% annotation}                                                                                                                                                                                                       & \multicolumn{4}{c}{20\% annotation}                                                                                                                                                                                                        \\ \cline{2-9} 
                             & \begin{tabular}[c]{@{}c@{}}Dice  (\%)↑\end{tabular} & \begin{tabular}[c]{@{}c@{}}Jaccard  (\%)↑\end{tabular} & \begin{tabular}[c]{@{}c@{}}95HD (voxel)↓\end{tabular} & \begin{tabular}[c]{@{}c@{}}ASD  (voxel)↓\end{tabular} & \begin{tabular}[c]{@{}c@{}}Dice  (\%)↑\end{tabular} & \begin{tabular}[c]{@{}c@{}}Jaccard  (\%)↑\end{tabular} & \begin{tabular}[c]{@{}c@{}}95HD  (voxel)↓\end{tabular} & \begin{tabular}[c]{@{}c@{}}ASD  (voxel)↓\end{tabular} \\ \hline
Class-agnostic feature consistency          & 90.55                                                 & 82.81                                                    & 5.94                                                     & 1.70                                                    & 91. 06                                                 & 83.73                                                    & 5.83                                                     & 1.49                                                    \\
Class-specific prototype alignment  & 91.81                                                 & 84.93                                                    & 5.08                                                     & 1.54                                                    & 92.11                                                 & 85.42                                                    & 4.98                                                     & 1.36                                                    \\ \hline
\end{tabular}}
\end{table*}

%% file: results/ablation_selfensembling.tex
\begin{table*}[!h]
    \centering\caption{Comparison between different self-ensembling strategies applied to our method, experimented on LA dataset with 10\% and 20\% labeled training data.}
    \label{tab:self-ensembling}
    \setlength{\tabcolsep}{2. mm}
    \resizebox{\linewidth}{!}{
\color{black}\begin{tabular}{c|cccc|cccc}
\hline
\multirow{2}{*}{Self-ensembling} & \multicolumn{4}{c|}{10\% annotation}     & \multicolumn{4}{c}{20\% annotation}      \\ \cline{2-9} 
     & \begin{tabular}[c]{@{}c@{}}Dice  (\%)↑\end{tabular} & \begin{tabular}[c]{@{}c@{}}Jaccard  (\%)↑\end{tabular} & \begin{tabular}[c]{@{}c@{}}95HD  (voxel)↓\end{tabular} & \begin{tabular}[c]{@{}c@{}}ASD  (voxel)↓\end{tabular} & \begin{tabular}[c]{@{}c@{}}Dice  (\%)↑\end{tabular} & \begin{tabular}[c]{@{}c@{}}Jaccard  (\%)↑\end{tabular} & \begin{tabular}[c]{@{}c@{}}95HD  (voxel)↓\end{tabular} & \begin{tabular}[c]{@{}c@{}}ASD  (voxel)↓\end{tabular} \\ \hline
Temporal Ensembling~\cite{laine2016temporal}          & 89.31    & 80.81     & 10.27       & 2.30       & 90.40    & 82.57       & 7.18       & 2.12      \\
Mean Teacher~\cite{tarvainen2017mean}                 & 91.81   & 84.93    & 5.08       & 1.54      & 92.11   & 85.42      & 4.98       & 1.36      \\ \hline
\end{tabular}}
\end{table*}

%% file: results/MixCompare.tex
\begin{table}[!h]
    \centering\caption{Comparison of segmentation performance between different mixing strategies used for our method.}
    \label{tab:mix}
    \setlength{\tabcolsep}{1.5 mm}
    \resizebox{0.999\linewidth}{!}{%
    \begin{tabular}{c|cccc|cccc}
        \hline
        \multirow{2}{*}{Method} & \multicolumn{4}{c|}{LA (10\%)}         & \multicolumn{4}{c}{Pancreas (10\%)}\\ \cline{2-9} 
     & \begin{tabular}[c]{@{}c@{}}Dice\\ (\%)↑\end{tabular} & \begin{tabular}[c]{@{}c@{}}Jaccard\\ (\%)↑\end{tabular} & \begin{tabular}[c]{@{}c@{}}95HD\\ (voxel)↓\end{tabular} & \begin{tabular}[c]{@{}c@{}}ASD\\ (voxel)↓\end{tabular} & \begin{tabular}[c]{@{}c@{}}Dice\\ (\%)↑\end{tabular} & \begin{tabular}[c]{@{}c@{}}Jaccard\\ (\%)↑\end{tabular} & \begin{tabular}[c]{@{}c@{}}95HD\\ (voxel)↓\end{tabular} & \begin{tabular}[c]{@{}c@{}}ASD\\ (voxel)↓\end{tabular} \\ \hline
        CutOut~\cite{devries2017improved}        & 90.73    & 83.10& 5.09 & 1.78   & 78.69  & 65.45& 8.96 & 3.28\\ 
        CutMix~\cite{yun2019cutmix}              & 90.85    & 83.32& 5.23 & 1.85   & 79.06  & 65.86& \textbf{7.04} & 2.30\\ 
        Copy-Paste~\cite{ghiasi2021simple}       & 91.06    & 83.64& 5.58 & 1.67   & 79.81  & 67.13& 7.29 & 1.72\\ 
        % B-Copy-Paste~\cite{bai2023bidirectional} & 91.34    & 84.14& 5.25 & 1.57   & 80.07  & 67.56& 7.89 & 2.10\\ 
        Mixup (ours)              & \textbf{91.81}    & \textbf{84.93} & \textbf{5.08} & \textbf{1.54}   & \textbf{80.72}  & \textbf{68.12} & 7.20 & \textbf{1.53} \\ \hline
        \end{tabular}
    }
\end{table}